\documentclass{article}

\usepackage{microtype}
\usepackage{graphicx}
\usepackage{booktabs} 

\usepackage{subcaption}
\usepackage{float}
\usepackage{enumitem}

\usepackage{wrapfig}

\usepackage{multirow}

\newcommand{\colorline}[1]{%
    \hspace*{-\fboxsep}%
    \colorbox{gray!30}{%
        \parbox{\linewidth}{%
            #1%
        }%
    }%
}

\usepackage{xcolor, colortbl}
\definecolor{Gray}{gray}{0.8}

\usepackage{algorithm}
\usepackage[noend]{algorithmic}
\usepackage{hyperref}

\usepackage[accepted]{my_preprint}

\usepackage[toc,page]{appendix}

\usepackage{amsmath}
\usepackage{amssymb}
\usepackage{mathtools}
\usepackage{amsthm}

\usepackage[capitalize,noabbrev]{cleveref}

\theoremstyle{plain}
\newtheorem{theorem}{Theorem}
\newtheorem{corollary}{Corollary}

\usepackage[textsize=tiny]{todonotes}

\icmltitlerunning{CAST: Cluster-Aware Self-Training for Tabular Data via Reliable Confidence}

\begin{document}

\twocolumn[
\icmltitle{CAST: Cluster-Aware Self-Training for Tabular Data via Reliable Confidence}




\begin{icmlauthorlist}
\icmlauthor{Minwook Kim}{cse}
\icmlauthor{Juseong Kim}{cse}
\icmlauthor{Ki Beom Kim}{cse}
\icmlauthor{Giltae Song}{cse,air}
\end{icmlauthorlist}

\icmlaffiliation{cse}{School of Computer Science and Engineering, Pusan National University, Busan, Republic of Korea }
\icmlaffiliation{air}{Center for Artificial Intelligence Research, Pusan National University, Busan, Republic of Korea}

\icmlcorrespondingauthor{Minwook Kim}{kmiiiaa@pusan.ac.kr}
\icmlcorrespondingauthor{Giltae Song}{gsong@pusan.ac.kr}

\icmlkeywords{Machine Learning, ICML}

\vskip 0.3in
]



\printAffiliationsAndNotice{}  

\begin{abstract}
Tabular data is one of the most widely used data modalities, encompassing numerous datasets with substantial amounts of unlabeled data.
Despite this prevalence, there is a notable lack of simple and versatile methods for utilizing unlabeled data in the tabular domain, where both gradient-boosting decision trees and neural networks are employed.
In this context, self-training has gained attraction due to its simplicity and versatility, yet it is vulnerable to noisy pseudo-labels caused by erroneous confidence.
Several solutions have been proposed to handle this problem, but they often compromise the inherent advantages of self-training, resulting in limited applicability in the tabular domain.
To address this issue, we explore a novel direction of \emph{reliable confidence in self-training contexts} and conclude that \emph{self-training can be improved by making that the confidence, which represents the value of the pseudo-label, aligns with the cluster assumption.}
In this regard, we propose \textbf{C}luster-\textbf{A}ware \textbf{S}elf-\textbf{T}raining (CAST) for tabular data, which enhances existing self-training algorithms at a negligible cost while maintaining simplicity and versatility. 
Concretely, CAST calibrates confidence by regularizing the classifier's confidence based on local density for each class in the labeled training data, resulting in lower confidence for pseudo-labels in low-density regions.
Extensive empirical evaluations on up to 21 real-world datasets confirm not only the superior performance of CAST but also its robustness in various setups in self-training contexts.
\end{abstract}

\section{Introduction} \label{sec:Introduction}
Tabular data is one of the most prevalent data modalities in real-world applications, often comprising a significant number of unlabeled samples.
However, handling these unlabeled samples is more challenging compared to other data types, such as those in computer vision or natural language processing.
While neural networks are the dominant architecture in many domains, they are not the only optimal solution in the tabular domain. 
Specifically, practitioners working with tabular data frequently use not only neural networks but also gradient-boosting decision trees (GBDTs) \cite{kaggle_2021_state, tabular_dnn_survey, dnn_is_not_all_u_need}.
Therefore, any approach that leverages the rich information in unlabeled samples should be simple and versatile, ensuring compatibility with both neural networks and GBDTs.
However, recent semi- and self-supervised learning methods for tabular data are not compatible with GBDTs, resulting in limited applicability \cite{vime, subtab, scarf, switchtab}.

In this context, self-training has gained attraction due to its simplicity and versatility in handling large amounts of unlabeled samples within the tabular domain, as it employs the identical training procedure used in supervised learning.
The key aspect of self-training is its iterative process: in each iteration, the classifier is trained with labeled data and then assigns pseudo-labels to unlabeled data based on its current predictions.
These pseudo-labeled samples are then used as the labeled samples in the subsequent iteration, gradually improving the classifier's performance.
Therefore, it is particularly useful for practitioners in the tabular domain, as it does not depend on a specific model architecture or training procedure.

Contemporary self-training methods consider the confidence, often referred to as prediction probabilities of the classifier, as the score and generate a pseudo-label if the confidence score is higher than or equal to a certain threshold \cite{pseudo_label3, pseudo_label4}.
However, it may not consistently serve as a reliable metric in real-world scenarios for various reasons such as biased classifiers or overconfidence in neural networks \cite{calibration1}.
These erroneous confidence scores can lead to the generation of noisy pseudo-labels during the self-training iterations, which may introduce confirmation bias that undermines the final self-training performance \cite{confirmation_bias}. 
Given these potential pitfalls, relying solely on the confidence may be a precarious choice \cite{confidence_regularized_self-training, uncertainty_pseudo-label, neighborhood-regularized}.

\begin{figure*}[tbp]
    \centering
        \centering
        \includegraphics[width=.8\linewidth]{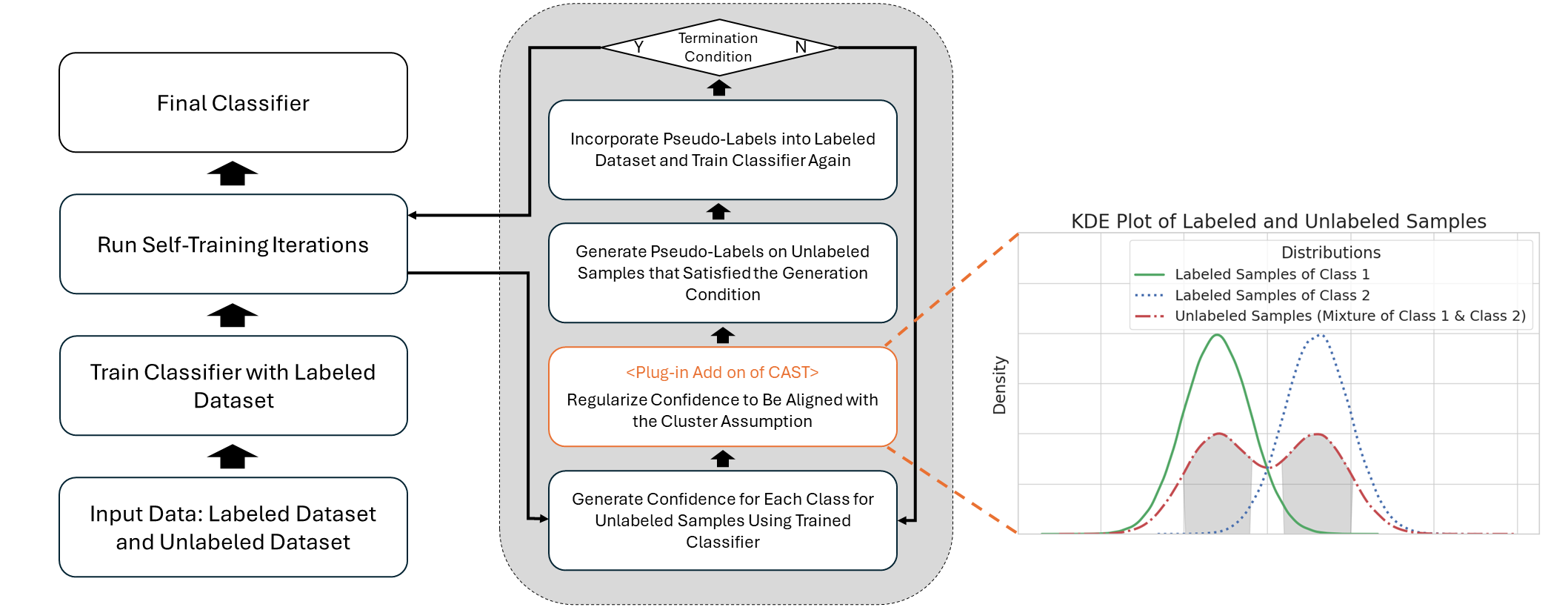}
        \label{fig:framework}
    \caption{Workflow of CAST. 
    CAST enhances self-training algorithms with a simple plug-in add-on.
    Unlike naive confidence-based methods that assign equal weights to all unlabeled samples, CAST prioritizes samples in high-density regions (shaded areas) over those in low-density regions. 
    This approach generates pseudo-labels that better align with the cluster assumption, thereby improving the effectiveness of self-training algorithms.
    }
\end{figure*}

Several studies have been conducted to improve erroneous confidence by calibrating the confidence to reflect its ground truth correctness likelihood \cite{calibration1}.
However, these solutions focus primarily on decision-making and are not verified in self-training contexts.
The current methods sought to counteract erroneous confidence during the self-training iterations by modifying the self-training algorithms or the model architectures \cite{setred, distance_editing, uncertainty_pseudo-label, reference-guided}.
Yet, these solutions often introduce significant overhead due to their modifications, and most are incompatible with GBDTs \cite{setred, distance_editing, uncertainty_pseudo-label, reference-guided}, which diminishes the advantages of self-training: simplicity and versatility.
For practitioners who want to apply reliable pseudo-labeling for self-training on tabular data, these are significant impediments.
Therefore, we raise a natural but ignored question: \emph{Can we improve self-training for tabular data by making confidence more reliable, without modifying the self-training algorithm or model architecture?}

After dissecting self-training, we argue that \emph{cluster assumption}, foundational to semi-supervised learning, can guide to reliable confidence in self-training contexts and conclude that the pseudo-labels that lie in high-density regions are more reliable than those that lie in low-density regions.
The cluster assumption states that data samples form clusters according to each class.
As such, the decision boundary should avoid high-density regions, favoring low-density regions instead \cite{low_density, modified_cluster_assumption, pseudo_label}.
Therefore, by assigning high confidence to pseudo-labels in high-density regions and low confidence to those in low-density regions, the self-training algorithm ensures that reliable pseudo-labels, which are far from the decision boundary, have higher confidence and are consistent with the cluster assumption.
In this regard, we propose \textbf{CAST}: \textbf{C}luster-\textbf{A}ware \textbf{S}elf-\textbf{T}raining for tabular data.
CAST calibrates confidence during the pseudo-labeling procedure by regularizing the classifier's confidence based on local density for each class in the labeled training data, resulting in lower confidence for pseudo-labels in low-density regions.
Consequently, CAST prioritizes pseudo-labels that are in high-density regions over those that are in low-density regions.
Figure 1 illustrates the workflow of CAST.

Our key contributions are summarized as follows: 
(1) We explore a novel direction of \emph{reliable confidence in self-training contexts} to address noisy pseudo-label issues. 
We conclude, both theoretically and empirically, that confidence in self-training should align with the cluster assumption, a foundational principle in semi-supervised learning.
(2) We propose a simple yet effective self-training enhancement algorithm, CAST, for tabular data.
CAST calibrates confidence based on the local density of unlabeled samples, resulting in reliable confidence that aligns with the cluster assumption.
(3) Our extensive experiments on up to 21 real-world classification datasets confirm that calibrated confidence of CAST consistently delivers marked performance enhancements across various setups at a negligible cost, while remaining orthogonal to existing algorithms.
More importantly, CAST extracts additional information from unlabeled samples, enhancing state-of-the-art semi- and self-supervised learning methods in the tabular domain.

\section{Related Work}

\textbf{Confidence Calibration.}
The erroneous confidence is one of the most prevalent problems in AI fields.
The primary methods to solve the issue is to calibrate the confidence for safe decision-making \cite{tree_calibration, calibration1, gnn_calibration}.
\citet{calibration1} define that a classifier is well-calibrated when its confidence estimates are representative of the true correctness likelihood.
This definition has been widely accepted across various studies \cite{ref_calibration3, gp_calibration, spline_calibration, ref_calibration1, ref_calibration2}.
One of the most widely used metrics for calibration to measure how well the classifier is calibrated is Expected Calibration Error (ECE)\footnote{Refer to the Appendix for more details about ECE.} \cite{ece}.
There are two primary strategies for achieving a well-calibrated model that produces reliable confidence. 
The first approach aims to calibrate the classifier during training \cite{ref_calibration3, ref_calibration1, ref_calibration2}, whereas the second performs post-hoc calibration by transforming the confidence of a given classifier \cite{gp_calibration, spline_calibration}.
Nevertheless, the impact of ECE-based calibration on self-training algorithms has not yet been investigated, despite the fact that confidence is one of the most crucial components in self-training.
To the best of our knowledge, this is the first work to investigate reliable confidence in the context of self-training.

\textbf{Recent Semi- and Self-Supervised Learning Methods for Tabular Data.}
Several semi- and self-supervised learning methods have been proposed to leverage large amounts of unlabeled samples in tabular datasets.
The most common approach involves using autoencoder-based models to learn useful representations from the rich information in unlabeled samples
\cite{vime, subtab, scarf, switchtab}.
More specifically, they utilize several pretext tasks such as feature reconstruction \cite{vime, subtab, switchtab} and contrastive learning \cite{subtab, scarf}, or perform consistency regularization \cite{vime}.
However, these are designed for neural networks, which limits their applicability in the broader context of tabular domains.

\textbf{Reliable Pseudo-Labeling for Self-Training.}
Reliable pseudo-labeling has attracted considerable interest in self-training contexts.
Early work focused on noise filtering to eliminate noisy pseudo-labels using 1) cut-edge weights \cite{setred, snnrce}; 2) distance of pseudo-labels \cite{selection-rejection, distance_editing}; 3) clustering analysis \cite{clustering_analysis}, but they often require excessive overhead, resulting in losing the simplicity of self-training.
Recent approaches have been slightly different, i.e. they do not use a noise filtering step during pseudo-labeling.
They have focused on generating pseudo-labels that are considered reliable pseudo-labels.
Specifically, these methods rely on the following: prototypes for pseudo-labels \cite{prototype, reference-guided}, stability of pseudo-labels \cite{st++}, uncertainty of pseudo-labels \cite{uncertainty_pseudo-label}, and neighborhood graph of pseudo-labels \cite{neighborhood-regularized}.
Additionally, some approaches enhance pseudo-labeling for self-training by using soft pseudo-labels based on confidence regularization \cite{confidence_regularized_self-training} or by introducing a debiased self-training framework \cite{debiased}.
However, these methods are not compatible with GBDTs, which restricts their use in tabular domains.

\section{Problem Statement} \label{sec:problem_statement}
Let, $C_{\theta}$ be a classifier, and $X_{U} = \{x^{(i)}\}_{i=1}^{U}$ be an unlabeled dataset with $U$ samples.
For a given $i^{th}$ unlabeled sample $\mathbf{x}^{(i)} \in X_{U}$, the pseudo-label $\Tilde{\mathbf{y}}^{(i)} = \begin{bmatrix}\Tilde{y}_{1}, \Tilde{y}_{2}, ... , \Tilde{y}_{N - 1}, \Tilde{y}_{N}\end{bmatrix}$ for an $N$-class dataset is generated during the self-training iterations based on a specified threshold $\tau$ and confidence $\mathbf{c} = \begin{bmatrix}c_{1}, c_{2}, ... , c_{N - 1}, c_{N}\end{bmatrix} \leftarrow C_{\theta}(\mathbf{x}^{(i)})$.
Formally,
\begin{equation} \label{eq:pseudo-labeling}
    \Tilde{y}_{j} = 
    \begin{Bmatrix*}[l]
    1 \quad \text{if} \ j \text{ = argmax}(\mathbf{c}) \text{ and } \text{max}(\mathbf{c}) \geq \tau \\
    0 \quad \text{otherwise}\\
  \end{Bmatrix*}
  .
\end{equation}  
In this paper, we aim to make confidence more reliable in self-training contexts without significant modifications, thereby simply improving self-training performance regardless of 
whether $C_{\theta}$ is a neural network or a gradient-boosting decision tree for practitioners working with tabular data.

\section{Method}\label{sec:method}
\subsection{Dissecting Self-Training}
\textbf{When are unlabeled samples informative in the semi-supervised learning settings?} 
Several parametric statistical studies have illustrated the value of unlabeled samples, whether for particular distributions \cite{specific_distribution} or for distribution mixtures \cite{mixture_distribution}.
These studies conclude that the information of unlabeled samples diminishes as class overlap increases \cite{unlabeled3, entropy_regularization}.
Therefore, semi-supervised learning typically relies on the cluster assumption, which postulates that data samples form clusters according to each class and decision boundaries should not cross high-density regions but instead lie in low-density regions \cite{low_density, unlabeled_data, ssl_survey, ssl_overview}.

\begin{figure}[tbp]
    \centering
    \begin{subfigure}[b]{0.35\linewidth}
        \centering
        \includegraphics[width=.95\linewidth]{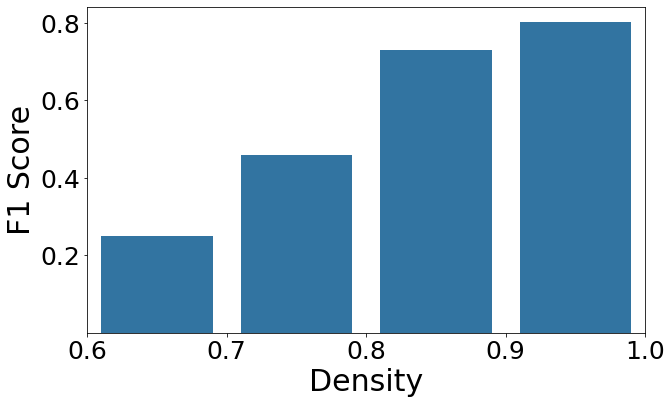}
        \caption{6M mortality}
        \label{fig:6m_per_density}
    \end{subfigure}
    \begin{subfigure}[b]{0.35\linewidth}
    \centering
        \includegraphics[width=.95\linewidth]{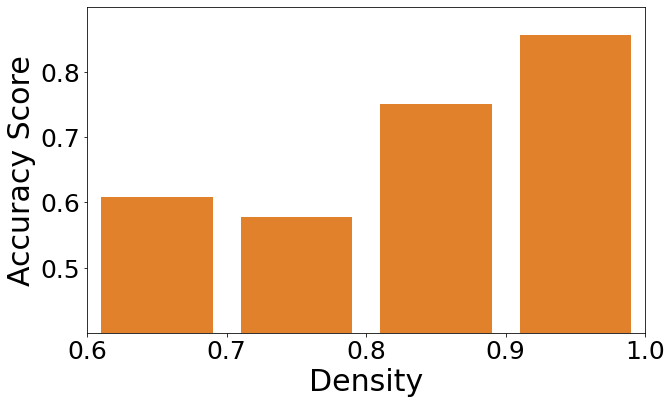}
        \caption{diabetes}
        \label{fig:diabetes_per_density}
        \end{subfigure}
    \\
    \begin{subfigure}[b]{0.35\linewidth}
        \centering
        \includegraphics[width=.95\linewidth]{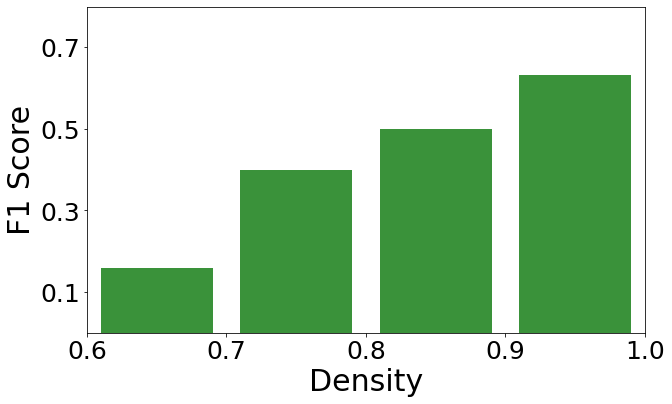}
        \caption{ozone}
        \label{fig:ozone_per_density}
    \end{subfigure}
    \begin{subfigure}[b]{0.35\linewidth}
        \centering
        \includegraphics[width=.95\linewidth]{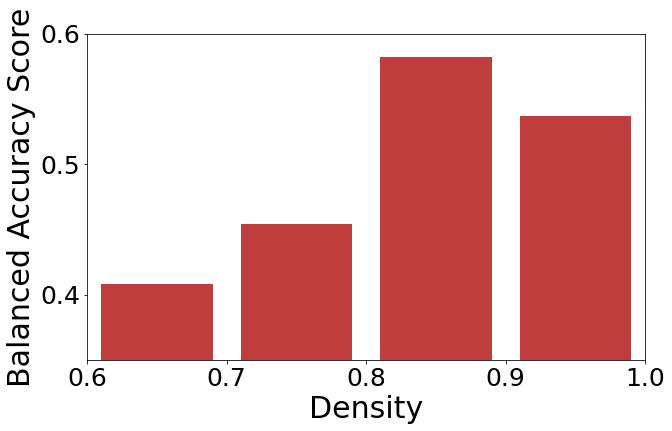}
        \caption{cmc}
        \label{fig:cmc_per_density}
    \end{subfigure}
    \caption{Evaluation results on pseudo-labels according to the density for each dataset.
    The pseudo-labels are produced by XGBoost \cite{xgboost}, while the empirical likelihood is used to estimate density.}
    \label{fig:per_density}
\end{figure}

\textbf{The objective of self-training and its limitation.}
Self-training is a version of the entropy minimization algorithm which minimizes the likelihood deprived of the entropy of the partition \cite{self-training_em}.
It constructs hard labels from high-confidence predictions on unlabeled data to implicitly achieve entropy minimization \cite{mixmatch}.
These techniques aim for the classifier to learn the low-density separations within the data, hence they assume that the given dataset satisfies the cluster assumption.
However, unreliable pseudo-labels that lie in low-density regions, stemming from erroneous confidence, violate the assumption and consequently disrupt the classifier's ability to learn the separations among classes during the self-training iterations. 
The empirical results in Figure \ref{fig:per_density} verify that the pseudo-labels in high-density regions are generally more reliable than those in low-density regions.

\subsection{Theoretical Motivation}\label{subsec:theoretical_motivation}
Here, we discuss the value of unlabeled samples based on their density by extending the theory of \citet{mixture_distribution} to the Corollary \ref{cor:corollary}.
For simplicity, we analyze a one-vs-one, binary classification task, but this analysis can be easily extended to a multiclass classification task by converting to a one-vs-rest task.
\begin{theorem}
\label{thm:theorem}
Let, $x$ be a unlabeled sample under the assumption that the densities $D$ of the observations for each class $y_1$, and $y_2$ are known.
The Fisher information, $I_{u}(\hat{p})$, for unlabeled samples at the estimate $\hat{p}$ is clearly a measure of the overlap between class conditional densities which denote the information content of unlabeled samples.
\begin{equation*}
    I_{u}(\hat{p}) = \int\frac{(D(x|y_1) - D(x|y_2))^2}{\hat{p}D(x|y_1) + (1-\hat{p})D(x|y_2)}dx
\end{equation*}
\end{theorem}
The following corollary deduced from Theorem \ref{thm:theorem} confirms that unlabeled samples in high-density regions are more informative than those in low-density regions.
%
\begin{corollary} \label{cor:corollary}
The information content of the unlabeled samples that lie in high-density regions ($I^{high}_{u}(\hat{p})$) is greater than those that lie in low-density regions ($I^{low}_{u}(\hat{p})$), i.e., $I^{high}_{u}(\hat{p}) > I^{low}_{u}(\hat{p})$.
\end{corollary}
For the proof of the Corollary \ref{cor:corollary}, refer to the Appendix.
\subsection{Cluster-Aware Self-Training} \label{subsec:cast}
Inspired by the aforementioned analyses, we argue that pseudo-labels for self-training are more reliable in high-density regions than in low-density regions and that the confidence should be aware of the cluster assumption.
Therefore, we propose CAST for tabular data which lowers the confidence of pseudo-labels that are in low-density regions to focus on those in high-density regions to improve self-training.
Concretely, CAST calibrates confidence by regularizing the confidence using local density for each class which is based on prior knowledge derived from labeled data.

Given $i^{th}$ unlabeled data $\mathbf{x}^{(i)}$, pseudo-label $\Tilde{\mathbf{y}}^{(i)} = \begin{bmatrix}\Tilde{y}_{1}, \Tilde{y}_{2}, ... , \Tilde{y}_{N - 1}, \Tilde{y}_{N}\end{bmatrix}$ for an $N$-class dataset is generated based on the confidence $\mathbf{c} = \begin{bmatrix}c_{1}, c_{2}, ... , c_{N - 1}, c_{N}\end{bmatrix} \leftarrow C_{\theta}(\mathbf{x}^{(i)})$, following the pseudo-labeling algorithm.
Typically, pseudo-labels are produced to the class with the highest confidence that exceeds a predefined threshold.
Since the pseudo-labels in high-density regions are more reliable, and the goal of self-training is to learn the low-density separations, we want to prioritize the pseudo-labels in high-density regions.
Therefore, we aim to regularize confidence to be aware of density that considers the cluster assumption, i.e., \emph{cluster-aware confidence}.

The natural characteristic of the tabular data is each feature occupies a specific, fixed position within the table.
This characteristic enables direct extraction of density using various parametric or nonparametric approaches based on observed data, i.e., the labeled training dataset. 
Hence, we utilize the prior knowledge for each class from the labeled training samples as the density estimation using a density estimator $D$.
However, prior knowledge derived from the training dataset is usually imperfect, particularly in semi-supervised learning settings where the labeled training data is scarce.
To address this issue, we extract the density only for the most important features, $\mathbf{\hat{x}}$, using feature selection.
Here, the prior knowledge $\boldsymbol{\gamma}$ for each class using $D_t$, which is fitted to the labeled training data distribution $t$, is defined as follows:
\begin{equation} \label{eq1} 
    \begin{aligned}
        \boldsymbol{\gamma}^{(i)} = \begin{bmatrix}\gamma_{1}, ... , \gamma_{N}\end{bmatrix}, 
        \quad \text{where} \,\, \gamma_{j} \leftarrow \textit{D}_t(\mathbf{\hat{x}}^{(i)}|y_j = 1)
    \end{aligned}
\end{equation}
Then, the regularized confidence that is aware of the cluster assumption is achieved by taking the element-wise product with prior knowledge $\boldsymbol{\gamma}$ and confidence $\mathbf{c}$.
\begin{equation}\label{eq2}
    \boldsymbol{\gamma} \circ \mathbf{c}
\end{equation} 
%
%
%
%
However, the use of naive $\boldsymbol{\gamma}$ in eq (\ref{eq2}) is incompatible with the typical threshold in existing pseudo-labeling algorithms, as $\boldsymbol{\gamma}$ is generally a low value, close to zero, especially for higher-dimensional datasets.
Therefore, we scale the $\boldsymbol{\gamma}$ using a min-max scaler before applying it to eq (\ref{eq2}).
Finally, in order to regulate the influence of possibly imperfect prior knowledge on the confidence, we adjust the balance between eq (\ref{eq2}) and $\mathbf{c}$ using the hyperparameter $\alpha$.
Our regularized confidence, $\mathbf{c}_r$, is defined as follows:
\begin{equation} \label{eq3}
    \mathbf{c}_r = \alpha (\boldsymbol{\gamma} \circ \mathbf{c}) + (1 - \alpha) \mathbf{c}
\end{equation}
In this eq (\ref{eq3}), the hyperparameter $\alpha$ delineates the influence of $\boldsymbol{\gamma}$ on regularizing the confidence.
If $\alpha$ is close to 0, the regularized confidence is close to the naive confidence, which is the same confidence used in conventional self-training.
Conversely, a high $\alpha$ value, approaching 1, steers the $\mathbf{c}_r$ to prioritize $\boldsymbol{\gamma} \circ \mathbf{c}$.
We have designed CAST to be adaptable, leaving the selection of the density estimator for $\boldsymbol{\gamma}$ to implementation, as the distribution of the data may vary and we believe that open to extension is a crucial component for versatility.
CAST is defined as any self-training algorithm that employs this regularized confidence $\mathbf{c}_r$ instead of the naive confidence.
The overall pseudo-code of CAST is shown in the Appendix.

\section{Experimental Evaluation}
In this section, we conduct extensive experiments to confirm the effectiveness of CAST in self-training contexts for tabular domains.
We evaluate its performance using various model architectures and different self-training setups to ensure robustness in the tabular domain.

This section is divided into four parts\footnote{In addition to these, we have included several additional experiments in the Appendix, including a demonstration of CAST's negligible computational cost.}. 
In the first part, we provide detailed information about the experiments, including the datasets, implementation details, and other relevant information.
In the second part, we evaluate the performance of our proposed method, CAST, under various setups in self-training contexts. 
In the third part, we visualize the impact of our method on the confidence of pseudo-labels using a toy dataset. 
Finally, we assess the significance of the hyperparameter $\alpha$ and the feature selection in CAST.
\begin{table*}[t]
  \small
  \setlength{\tabcolsep}{1.5mm}
  \centering
  \begin{tabular}{lccccccccccccc}
    \toprule
    & & \multicolumn{3}{c}{6M mortality} & \multicolumn{3}{c}{diabetes} & \multicolumn{3}{c}{ozone} & \multicolumn{3}{c}{cmc} \\ 
    \cmidrule(r){3-5}
    \cmidrule(r){6-8}
    \cmidrule(r){9-11}
    \cmidrule(r){12-14}
        &     &      XGB   &    FT & MLP & XGB & FT & MLP & XGB & FT & MLP & XGB & FT & MLP \\
    \midrule
    \multicolumn{2}{c}{w/o ST} & 0.4055  &    0.3806   &    0.3311   &    0.6613   &    0.7143   &    0.7152   &    0.2803   &    0.3769   &    0.3823   &    0.4638   &    0.4696   &    0.4437 \\
    \midrule
    \multirow{7}{*}{FPL} & Baseline &    0.4221  &    0.3849   &    0.3571   &    0.6613   &    0.7167   &    0.7245   &    0.2813   &    0.3781   &    0.3872   &    0.4674   &    0.4708   &    0.4443  \\
    & TS &    0.4221 & 0.3849 & 0.3566 & 0.6613 & 0.7167 & 0.7245 & 0.2813 & 0.3782 & 0.3872 & 0.4674 & 0.4708 & 0.4443 \\
    & HB &    0.4222 & 0.3806 & 0.3306 & 0.6615 & \underline{0.7214} & 0.7208 & 0.2787 & 0.3864 & 0.3817 & 0.4652 & 0.4745 & 0.4481 \\
    & SP & \underline{0.4228} & 0.3895 & 0.3590 & 0.6606 & 0.7158 & 0.7214 & 0.2793 & 0.3958 & 0.3921 & 0.4628 & \underline{0.4762} & 0.4437 \\
    & GP & 0.4099 & 0.3764 & 0.3308 & 0.6665 & 0.7199 & 0.7136 & 0.2834 & 0.3807 & 0.3768 & 0.4638 & 0.4696 & 0.4437 \\
    \rowcolor{Gray}
    \cellcolor{white} & CAST-D & 0.4221 & \underline{0.4018} & \underline{0.3660} & \textbf{0.6719} & \textbf{0.7242} & \underline{0.7275} & \textbf{0.3009} & \textbf{0.4103} & \textbf{0.4169} & \underline{0.4746} & 0.4730 & \textbf{0.4509} \\
    \rowcolor{Gray}
    \cellcolor{white} & CAST-L & \textbf{0.4444} & \textbf{0.4147} & \textbf{0.3873} & \underline{0.6701} & 0.7190 & \textbf{0.7292} & \underline{0.2988} & \underline{0.4022} & \underline{0.4131} & \textbf{0.4747} & \textbf{0.4780} & \underline{0.4483} \\
    \midrule
    \multirow{7}{*}{CPL} & Baseline & 0.4029 & 0.4039 & 0.3471 & 0.6621 & 0.7201 & 0.7208 & 0.2943 & 0.3683 & 0.4047 & 0.4677 & 0.4786 & 0.4456 \\
    & TS & 0.4029 & 0.4039 & 0.3471 & 0.6621 & 0.7201 & 0.7208 & 0.2943 & 0.3683 & 0.4038 & 0.4677 & 0.4786 & 0.4456 \\
    & HB & \underline{0.4252} & 0.3893 & 0.3536 & 0.6621 & 0.7260 & 0.7177 & 0.2820 & 0.3845 & 0.3840 & 0.4671 & 0.4751 & 0.4447 \\
    & SP & 0.4024 & 0.4047 & 0.3586 & 0.6602 & 0.7245 & 0.7180 & 0.2859 & 0.3817 & 0.4041 & 0.4641 & 0.4773 & 0.4520 \\
    & GP & 0.4048 & 0.4005 & 0.3449 & 0.6675 & 0.7173 & 0.7139 & 0.2952 & 0.3831 & 0.4001 & 0.4667 & 0.4764 & 0.4466 \\
    \rowcolor{Gray}
    \cellcolor{white} & CAST-D & 0.4066 & \underline{0.4162} & \underline{0.3694} & \textbf{0.6701} & \underline{0.7331} & \underline{0.7271} & \underline{0.2956} & \textbf{0.4176} & \textbf{0.4182} & \underline{0.4800} & \textbf{0.4919} & \textbf{0.4595} \\
    \rowcolor{Gray}
    \cellcolor{white} & CAST-L & \textbf{0.4359} & \textbf{0.4290} & \textbf{0.3897} & \underline{0.6677} & \textbf{0.7338} & \textbf{0.7364} & \textbf{0.3133} & \underline{0.4093} & \underline{0.4165} & \textbf{0.4811} & \underline{0.4896} & \underline{0.4595} \\
    \bottomrule
  \end{tabular}
  \caption{Performance comparison over four tabular datasets. The best results are highlighted in bold, while the second-best scores are underlined. Abbreviations are as follows: without self-training (w/o ST), temperature scaling (TS), histogram binning (HB), spline calibration (SP), and latent Gaussian process (GP).}
  \label{table:performance_comparison}
\end{table*}
\subsection{Experimental Details}
In this section, we provide a brief overview of the experimental details.
Further information can be found in the Appendix.

\textbf{Datasets.}
We mainly evaluate the various confidences in self-training contexts using four tabular datasets.
First, we adopt the 6-month mortality prediction after acute myocardial infarction (in short, 6M mortality) dataset from the Korea Acute Myocardial Infarction Registry (KAMIR).
The scarcity of labels in the dataset inspired us to study self-training in the tabular domain.
The other three datasets (diabetes, ozone, and cmc) are sourced from OpenML-CC18—a benchmark suite of meticulously curated datasets \citep{openml, openml_suites, openml_python}.
Our choice of these datasets aims to illustrate the impact of CAST across diverse tabular domains.
Unless otherwise noted, we report the F1 score for 6M mortality and ozone, the accuracy score for diabetes, and the balanced accuracy score for the cmc dataset, considering the impact of class imbalance on performance.
We also conduct extended empirical experiments using an additional seventeen datasets from OpenML-CC18 to demonstrate the results for broader domains, which are reported in the Appendix.

\textbf{Compared Methods.}
Our evaluations use two primary pseudo-labeling strategies for self-training: fixed-threshold pseudo-labeling (FPL) and curriculum pseudo-labeling (CPL). 
The former employs a fixed threshold \cite{combining, rethinking, noisy_student}, while the latter uses a dynamic threshold that is lowered during self-training iterations to generate pseudo-labels \cite{curriculum_labeling, flexmatch}.
Using these self-training methods, we establish baselines as naive confidence-based self-training methods.

Additionally, we compare the confidence of our proposed methods with other approaches using several confidence calibration techniques known to enhance reliability.
We select temperature scaling and histogram binning for the confidence calibration methods because of their simplicity and widespread use \cite{calibration1}.
Furthermore, we employ spline calibration \cite{spline_calibration} and latent Gaussian process calibration \cite{gp_calibration} to evaluate more sophisticated confidence calibration methods.

\textbf{Model Architectures.} 
To demonstrate the versatility of our proposed method, CAST, we use various model architectures in the tabular domain as base classifiers for self-training.
Specifically, we employ XGBoost \cite{xgboost}, FT-Transformer \cite{ft-transformer}, and MLP, representing GBDTs, transformer-based models, and basic neural networks, respectively.
Additionally, we utilize VIME \cite{vime}, SubTab \cite{subtab}, SCARF \cite{scarf}, and SwitchTab \cite{switchtab}, which are recent state-of-the-art semi- and self-supervised learning models in the tabular domain.

\textbf{Training and Evaluation Details.}  
Given that the ultimate goal of semi-supervised learning is to surpass the performance of well-tuned supervised models \cite{realistic_ssl}, we optimize each model using Optuna \cite{optuna} for 100 trials for supervised learning models and for 50 trials for semi- and self-supervised learning models.
In addition, as noted by previous studies \cite{realistic_ssl, realistic_ssl2}, relying solely on an insufficient validation set can lead to suboptimal hyperparameter selection.
Thus, we reserve 20\% of the data for the test dataset and employ 3-fold cross-validation on the remainder.
Unless otherwise noted, 10\% of the training dataset is randomly selected as labeled data, with the remainder serving as unlabeled data for self-training.

For FPL, we empirically adopt a threshold, $\tau$, of 0.6. 
For CPL, we set the starting threshold to capture the top 20\% and incrementally increase the percentage by 20\%, following the recommendations of Cascante-Bonilla et al. \citeyearpar{curriculum_labeling}. 
Self-training iterations are terminated under two conditions: for FPL, when a self-trained classifier underperforms after self-training iteration, and for CPL, when no additional unlabeled data remain.
To mitigate confirmation bias accumulation during self-training iterations, we reinitialize all classifiers after generating pseudo-labels, as recommended by Cascante-Bonilla et al. \citeyearpar{curriculum_labeling}. 

All evaluations are conducted using ten random seeds ranging from 0 to 9, and the results are averaged across these runs.

\textbf{Implementation Details of CAST.} 
We adopt a multivariate kernel density estimator and empirical likelihood as density estimators to derive prior knowledge for CAST, denoting them as CAST-D and CAST-L, respectively.
To determine the optimal $\alpha$ value for CAST, we execute a grid search in eight steps over the range [0.2, 0.75]

\subsection{Results and Discussion} \label{subsec:results_and_discussion}
\textbf{While existing confidence calibration methods show little to no distinction compared to naive confidence, CAST significantly enhances confidence for self-training.}
Here, we conduct a comprehensive performance comparison using various tabular models and datasets in self-training contexts, specifically FPL and CPL.
Intuitively, reliable confidence in the self-training context should yield superior performance compared with naive confidence.
However, as summarized in Table \ref{table:performance_comparison}, self-training approaches based on existing calibrated confidence methods often do not lead to performance improvement, and at times even diminish the final performance compared to self-training with naive confidence.
Contrarily, CAST consistently delivers notable enhancements in self-training across various strategies, datasets, and models.
In all conducted experiments, CAST often outperforms the other approaches, securing the top position in every experiment and ranking second in most.
The statistical analyses in the Appendix support the superiority of CAST.

\begin{figure*}[t]
    \centering
    \begin{subfigure}[b]{0.24\linewidth}
        \centering
        \includegraphics[width=.9\linewidth, height=0.09\textheight]{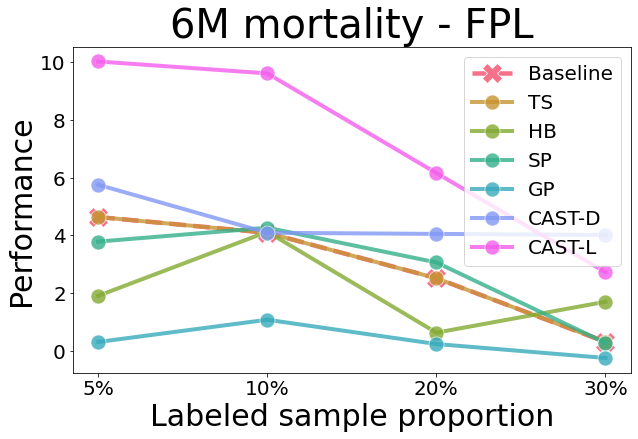}
        \label{fig:4-(a)}
    \end{subfigure}
    \begin{subfigure}[b]{0.24\linewidth}
    \centering
        \includegraphics[width=.9\linewidth, height=0.09\textheight]{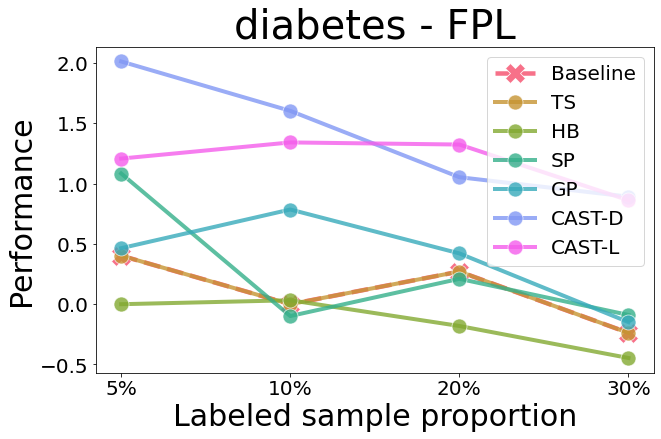}
        \label{fig:4-(b)}
        \end{subfigure}
    \begin{subfigure}[b]{0.24\linewidth}
        \centering
        \includegraphics[width=.9\linewidth, height=0.09\textheight]{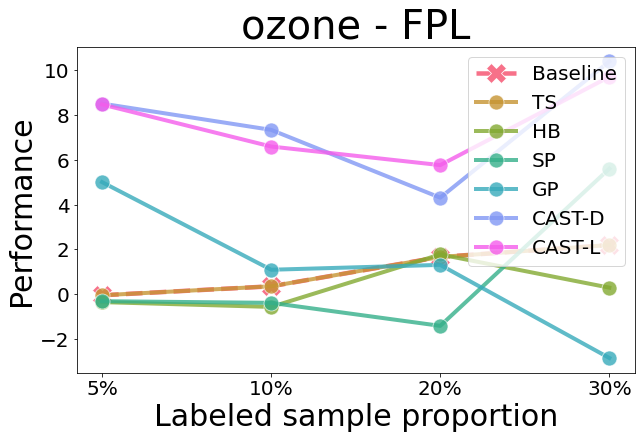}
        \label{fig:4-(c)}
    \end{subfigure}
    \begin{subfigure}[b]{0.24\linewidth}
        \centering
        \includegraphics[width=.9\linewidth, height=0.09\textheight]{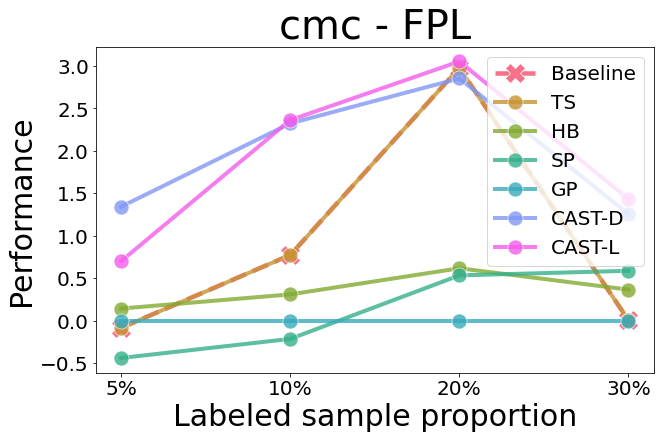}
        \label{fig:4-(d)}
    \end{subfigure}
    \\
    \begin{subfigure}[b]{0.24\linewidth}
        \centering
        \includegraphics[width=.9\linewidth, height=0.09\textheight]{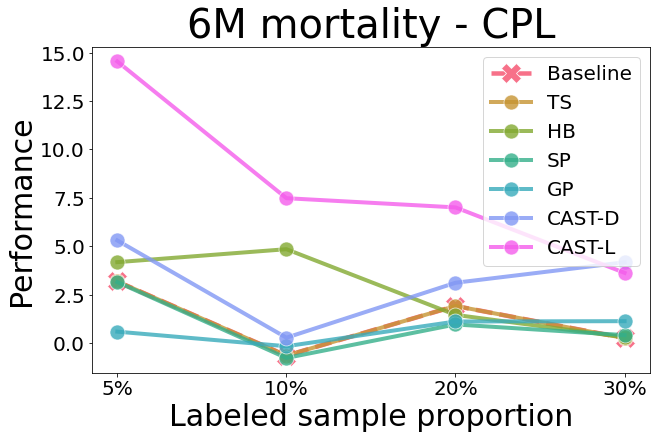}
        \label{fig:4-(e)}
    \end{subfigure}
    \begin{subfigure}[b]{0.24\linewidth}
    \centering
        \includegraphics[width=.9\linewidth, height=0.09\textheight]{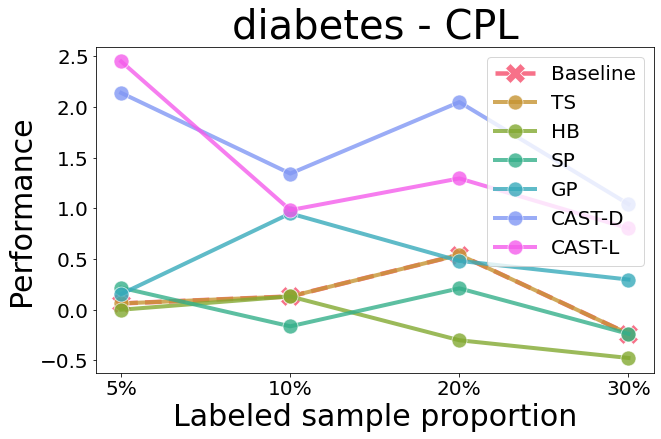}
        \label{fig:4-(f)}
        \end{subfigure}
    \begin{subfigure}[b]{0.24\linewidth}
        \centering
        \includegraphics[width=.9\linewidth, height=0.09\textheight]{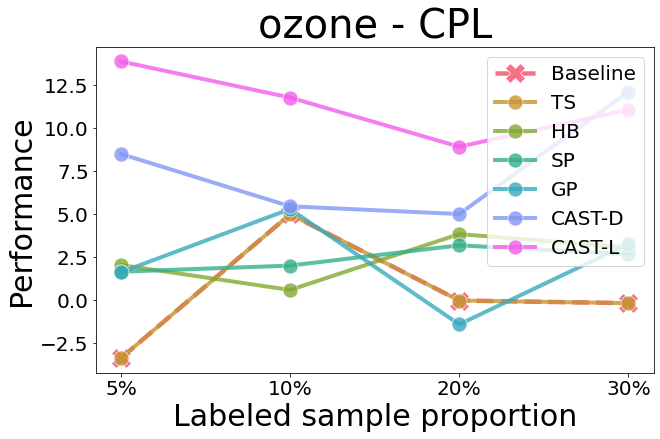}
        \label{fig:4-(g)}
    \end{subfigure}
    \begin{subfigure}[b]{0.24\linewidth}
        \centering
        \includegraphics[width=.9\linewidth, height=0.09\textheight]{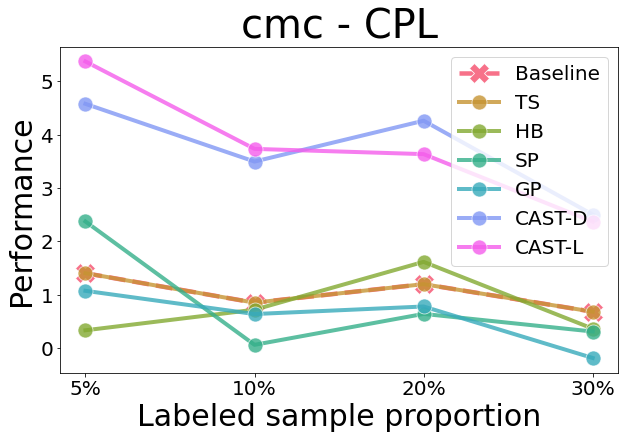}
        \label{fig:4-(h)}
    \end{subfigure}
    \caption{Relative improvement (\%) of various confidence-based self-training over various proportions of labeled samples in the training dataset. Abbreviations are as follows: temperature scaling (TS), histogram binning (HB), spline calibration (SP), and latent Gaussian process (GP).}
    \label{fig:various_sample_ratios}
\end{figure*}

\textbf{CAST demonstrates robustness for various labeled sample proportions.} 
Given that CAST derives prior knowledge from labeled data within the training dataset, we assess its effectiveness across various labeled sample proportions.
We present the results of self-training using different confidences at labeled training sample proportions of \{5\%, 10\%, 20\%, 30\%\} with XGBoost across the four datasets in Figure \ref{fig:various_sample_ratios}.
As illustrated in Figure \ref{fig:various_sample_ratios}, CAST consistently outperforms naive confidence-based self-training, irrespective of the labeled sample proportion in the training dataset.
These results underscore the robustness of CAST to variations in the proportion of labeled samples.

\begin{table}[t]
  \small
  \centering
  \setlength{\tabcolsep}{1.5mm}
  \begin{tabular}{lcccccc}
    \toprule
     & & 6M mortality & diabetes & ozone & cmc \\ 
    \midrule
    & w/o ST & 0.3788 & 0.7074 & 0.2146 & 0.4615 \\
    \midrule
    \multirow{7}{*}{FPL} & Baseline & 0.4035 & 0.7074 & 0.2081 & 0.4646 \\
                         & TS       & 0.3987 & 0.7076 & \textbf{0.2435} & 0.4646 \\
                         & HB       & 0.4052 & 0.7048 & 0.2365 & 0.4589 \\ 
                         & SP       & 0.4009 & 0.7065 & 0.2057 & 0.4637 \\ 
                         & GP       & 0.3715 & \textbf{0.7203} & 0.2255 & 0.4548 \\
                         \rowcolor{Gray}
                         & CAST-D   & \underline{0.4102} & 0.7180 & \underline{0.2412} & \underline{0.4670} \\ 
                         \rowcolor{Gray}
                         & CAST-L   & \textbf{0.4277 }& \underline{0.7195} & 0.2328 & \textbf{0.4675} \\ 
    \midrule
    \multirow{7}{*}{CPL} & Baseline & 0.3991 & 0.7084 & 0.2343 & 0.4666 \\
                         & TS       & 0.4010 & 0.7061 & 0.2179 & 0.4666 \\
                         & HB       & 0.4017 & 0.7050 & 0.2187 & 0.4638 \\ 
                         & SP       & 0.4002 & 0.7113 & 0.2187 & 0.4683 \\ 
                         & GP       & 0.3880 & 0.7078 & 0.2470 & 0.4590 \\
                         \rowcolor{Gray}
                         & CAST-D   & \underline{0.4136} & \underline{0.7165} & \underline{0.2503} & \underline{0.4793} \\ 
                         \rowcolor{Gray}
                         & CAST-L   & \textbf{0.4249} & \textbf{0.7236} & \textbf{0.2662} & \textbf{0.4808} \\ 
    \bottomrule
  \end{tabular}
  \caption{Performance comparison over four tabular datasets with corrupted features. The best results are highlighted in bold, while the second-best scores are underlined. Abbreviations are as follows: without self-training (w/o ST), temperature scaling (TS), histogram binning (HB), spline calibration (SP), and latent Gaussian process (GP).}
  \label{table:CAST_with_corrupted_features}
\end{table}

\textbf{CAST is robust to feature corruption.}
Feature corruption is a common problem in many real-world scenarios.
We investigate the effects of different confidences using XGBoost on datasets with corrupted features to demonstrate the robustness of CAST for noisy features.
%
We outline the methodology for inducing feature corruption as follows.
We randomly select a fraction of the features and replace each chosen feature with a value drawn from the empirical marginal distribution of that feature.
This distribution is defined as a uniform distribution over the values that the feature takes on across the training dataset.
The corruption ratio is fixed at 20\% for each training sample.
The results are summarized in Table \ref{table:CAST_with_corrupted_features}.
Clearly, CASTs consistently show notable performance improvements even in the presence of corrupted features.

\begin{table}[tbp]
    \centering
    \small
    \setlength{\tabcolsep}{1.0mm}
    \begin{tabular}{cccccc}
        \toprule
             & & 6M mortality  &  diabetes  & ozone & cmc \\
        \midrule
       \multirow{3}{*}{VIME} & w/o ST & 0.4432 & 0.7143 & 0.3908 & 0.4294 \\
       \cmidrule{2-6}
       & Baseline  &  0.4500 & 0.7165 & 0.3894 & 0.4402 \\
       &  CAST-L    & \textbf{0.4819} & \textbf{0.7249} & \textbf{0.4073} & \textbf{0.4403} \\
       \midrule
        \multirow{3}{*}{SubTab} & w/o ST & 0.4144 & 0.7110 & 0.3696 & 0.5008 \\
        \cmidrule{2-6}
        & Baseline & 0.4389 & 0.7143 & 0.3680 & 0.5078 \\
        &  CAST-L   & \textbf{0.4628} & \textbf{0.7268} & \textbf{0.4013} & \textbf{0.5253} \\
       \midrule
        \multirow{3}{*}{SCARF} & w/o ST & 0.2527 & 0.6972 & 0.2182 & 0.4765 \\
        \cmidrule{2-6}
        & Baseline & 0.2606 & 0.6994 & 0.2280 & 0.4973 \\
       &  CAST-L   & \textbf{0.3137} & \textbf{0.7076} & \textbf{0.2612} & \textbf{0.5190} \\
       \midrule
       \multirow{3}{*}{SwitchTab} & w/o ST & 0.4178 & 0.7056 & 0.3703 & 0.4723 \\
        \cmidrule{2-6}
       & Baseline & 0.4447 & 0.7193 & 0.3846 & 0.4819 \\
       & CAST-L & \textbf{0.4618} & \textbf{0.7333} & \textbf{0.4014} & \textbf{0.4957} \\
       \bottomrule
    \end{tabular}
    \caption{Performance comparison over four tabular datasets. ``w/o ST" denotes ``without self-training" and the best results are highlighted in bold.}
    \label{table:cast_with_ssl}
\end{table}

\textbf{CAST elevates state-of-the-art semi- and self- supervised learning tabular models to superior performance.}
Here, we compare naive confidence based self-training and CAST-L within FPL strategy using state-of-the-art semi- and self-supervised learning tabular models.
The results are shown in Table \ref{table:cast_with_ssl}.
As shown in Table \ref{table:cast_with_ssl}, self-training based on naive-confidence provides marginal improvements and sometimes makes performance worse.
However, CAST-L derives additional information from unlabeled samples, leading to greater performance gains for SOTA models.

\subsection{Visualization and Discussion}
\begin{figure}[tbp]
    \centering
    \begin{subfigure}[b]{0.45\linewidth}
    \centering
        \includegraphics[width=.95\linewidth]{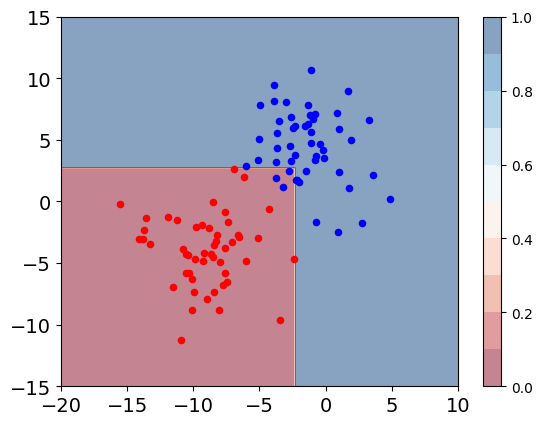}
        \label{fig:naive_confidence}
        \end{subfigure}
    \begin{subfigure}[b]{0.45\linewidth}
        \centering
        \includegraphics[width=.95\linewidth]{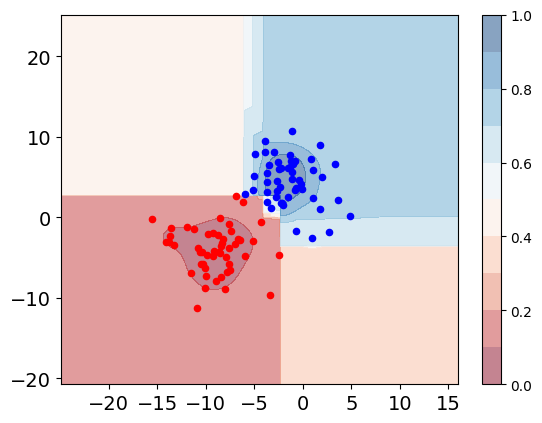}
        \label{fig:cast_confidence}
    \end{subfigure}
    \caption{Visualization of the confidence levels of XGBoost on the Blob dataset when generating pseudo-labels for the self-training using (Left) naive confidence and (Right) CAST-L. Colored points represent labeled samples in the training dataset for each class, and the degree of the color indicates the confidence level in the space where the Blob data exists.}
\label{fig:confidence_level}
\end{figure}

In this section, we visualize the confidence levels of XGBoost on unlabeled samples during self-training iterations using a binary classification toy dataset, Blob.
As shown in Figure \ref{fig:confidence_level}, the confidence levels of CAST-L exhibit lower values for samples in low-density regions, which might be unreliable for self-training.
In contrast, the confidence levels of naive confidence do not differentiate between high and low-density regions.

\subsection{Ablation Study}\label{subsec:ablation_study}
In this section, we present an ablation study to evaluate the significance of the hyperparameter $\alpha$ and the feature selection in CAST. 
For comparison, we employ XGBoost and FPL.
Table \ref{table:ablation} provides a summary of the results from our ablation study. 
A key observation is the pivotal role of the hyperparameter $\alpha$ in driving performance. 
The results indicate that CAST, when operating without $\alpha$ (thus generating $\mathbf{c}_r$ using eq. (\ref{eq2})), exhibits inferior performance compared to configurations that include $\alpha$.
In some cases, it even underperforms the baseline that relies on naive confidence for pseudo-labeling.
Further discussion on how to select $\alpha$ can be found in the Appendix.

Additionally, the study underscores the importance of feature selection.
The absence of feature selection often leads to suboptimal performance. 
This indicates that feature selection is critical for more accurate density estimation, which in turn leads to more precise density-based confidence regularization. 
To sum up, the hyperparameter $\alpha$ and the feature selection of CAST are critical components for proper confidence regularization.
\begin{table}[tbp]
    \centering
    \small
    \setlength{\tabcolsep}{1.0mm}
    \begin{tabular}{cccccc}
        \toprule
         & & 6M mortality  &  diabetes  & ozone & cmc \\
        \midrule
       Baseline  & & 0.4221 & 0.6613 & 0.2813 & 0.4674 \\
       \midrule
       \multirow{4}{*}{CAST-D} & w/o $\alpha$ \& fs & 0.4055 & 0.6647 & 0.2803 & 0.4658 \\
       & w/o $\alpha$ & 0.4061 & 0.6641 & 0.2739 & 0.4639 \\
       & w/o fs & \textbf{0.4223} & 0.6680 & 0.2852 & 0.4719 \\
         & with $\alpha$ \& fs & 0.4221 & \textbf{0.6719} & \textbf{0.3009} & \textbf{0.4746} \\
        \midrule
        \multirow{4}{*}{CAST-L} & w/o $\alpha$ \& fs & 0.4165 & 0.6604 & 0.2812 & 0.4651 \\
        & w/o $\alpha$ & 0.4327 & 0.6626 & 0.2837 & 0.4673 \\
        & w/o fs & 0.4319 & 0.6649 & 0.2953 & 0.4723 \\
         & with $\alpha$ \& fs & \textbf{0.4444} & \textbf{0.6701} & \textbf{0.2988} & \textbf{0.4747} \\
       \bottomrule
    \end{tabular}
    \caption{Ablation of $\alpha$ and feature selection. ``w/o" and ``fs" denote ``without" and ``feature selection," respectively. The best results are highlighted in bold.}
    \label{table:ablation}
\end{table}

\section{Conclusion}
Current efforts to resolve erroneous confidence during self-training in tabular domains have lost the simplicity and versatility of self-training, resulting in limited applicability to tabular practitioners.
This paper argues that self-training can be improved by making confidence more reliable without significant modifications for tabular practitioners, and identifies that the cluster assumption should be considered for reliable confidence in self-training contexts.
In this study, we propose a self-training enhancement algorithm, CAST, for tabular data that calibrates confidence to prioritize pseudo-labels that are in high-density regions over those that are in low-density regions without significant modifications to existing algorithms.
The extensive experiments across diverse settings verify the effectiveness of CAST in the tabular domain.

\bibliography{my_bibtex}
\bibliographystyle{my_preprint}

\clearpage

\onecolumn

\begin{appendices}

\setcounter{table}{0}
\renewcommand{\thetable}{A\arabic{table}}

\setcounter{figure}{0}
\renewcommand\thefigure{A\arabic{figure}}

\section{Broader Impacts and Limitations} \label{sec:broader_impact_and_limitations}

Tabular data is the most common data type in real-world, and our work can shed light on developing better algorithms for tabular practitioners with vast amounts of unlabeled samples.
For the tabular domain, \emph{CAST can enhance any self-training algorithm at a negligible cost, regardless of the model architecture}.
In addition, this is \emph{the first work on reliable confidence in the self-training context}, to the best of our knowledge.

However, our work has several limitations.
Firstly, our finding that pseudo-labels in high-density regions are more reliable can be extended to other domains.
However, common density estimation methods for other domains, such as image or text, are based on generative models, and incorporating them with self-training iterations greatly diminishes the simplicity of self-training.
As a result, the applicability of CAST to other domains is limited.
Additionally, our work identifies the essential components for reliable confidence in self-training contexts, but we evaluate the confidence indirectly via the performance on the test dataset after self-training iterations.
There are no direct assessments of confidence in the context of self-training.
We leave these for future work.

\section{Pseudo Code of CAST} \label{sec:general_form}
\begin{algorithm}[htp]
   \caption{CAST} 
   \label{alg:algorithm1}
\begin{algorithmic}
    \STATE {\bfseries Input:} Labeled and unlabeled dataset $X_{L}$ and $X_{U}$; 
    pseudo-labeling algorithm $\Phi$; target classifier $C$; performance metric $P$, Density estimator $D_t$ which is fitted to the labeled training data distribution $t$; feature selector $fs$; hyperparameter $\alpha$.
    \STATE {\bfseries Output:} The best classifier during the self-training iterations, $C_{best}$.
   \newline
   \colorline{
   \STATE $\boldsymbol{\Gamma} \leftarrow \varnothing$ \hspace*{\fill} Caching the prior knowledge
   \FOR{$\mathbf{x}^{(i)}$ in $X_{U}$}
        \STATE \quad $\hat{\mathbf{x}}^{(i)} \leftarrow fs(\mathbf{x}^{(i)})$
        \STATE \quad $\boldsymbol{\gamma}^{(i)} \leftarrow D_{t}(\hat{\mathbf{x}}^{(i)}), \quad \text{where} \quad \boldsymbol{\gamma}^{(i)} = \begin{bmatrix}\gamma_{1}, \gamma_{2}, ... , \gamma_{N - 1}, \gamma_{N}\end{bmatrix}$
        \STATE \quad $\boldsymbol{\Gamma} \leftarrow \boldsymbol{\Gamma} \cup \boldsymbol{\gamma}^{(i)}$ 
    \ENDFOR
    \STATE $\boldsymbol{\Gamma} \leftarrow$ Min-Max Scaler$(\boldsymbol{\Gamma})$ \hspace*{\fill} Plug-in of CAST
    }
   \STATE $C_{current} \leftarrow$ trained classifier on $X_{L}$
   \STATE $C_{best} \leftarrow C_{current}$
    
   \WHILE{the termination conditions of $\Phi$ are not met} 
   \STATE $\Tilde{X} \leftarrow X_{L}$
   \FOR{$i, \, \mathbf{x}^{(i)}$ in enumerate($X_{U}$)}
        \STATE $\mathbf{c} \leftarrow C_{current}(\mathbf{x}^{(i)})$
        
        \colorline{%
            \STATE $\boldsymbol{\gamma}^{(i)} \leftarrow \boldsymbol{\Gamma}(i)$ \hspace*{\fill}  Replace $\mathbf{c}$ with $\mathbf{c}_r$
            \STATE $\mathbf{c} \leftarrow \alpha (\boldsymbol{\gamma^{(i)}} \circ \mathbf{c}) + (1 - \alpha) \mathbf{c}$ \hspace*{\fill}  Plug-in of CAST
        }
        \STATE $\Tilde{\mathbf{y}}^{(i)} \leftarrow \Phi(\mathbf{c})$ \text{ // if } $\Phi$ \text{assigns a pseudo-label, then } $\Tilde{\mathbf{y}}^{(i)}$ \text{is a one-hot vector, otherwise a zero vector.}
        \IF {$\Tilde{\mathbf{y}}^{(i)} \neq \vec{0}$}
            \STATE $\Tilde{X} \leftarrow \Tilde{X} \cup \{(\mathbf{x}^{(i)}, \Tilde{\mathbf{y}}^{(i)})\}$
        \ENDIF
   \ENDFOR

   \STATE $C_{current} \leftarrow$ a classifier newly trained on $\Tilde{X}$
   \IF{$P(C_{current}) > P(C_{best})$}
        \STATE $C_{best} \leftarrow C_{current}$
   \ENDIF
   \ENDWHILE
   \STATE {\bfseries Return:} $C_{best}$
\end{algorithmic}
\end{algorithm}

\clearpage

\section{Expected Calibration Error (ECE)} \label{sec:ece}

The Expected Calibration Error (ECE) \citep{ece} quantifies the discrepancy between a model's predicted confidence and its true accuracy. To compute the ECE, predictions are grouped into $M$ bins of equal sizes based on their confidence, and the difference between the average accuracy and average confidence for each bin is determined.

Formally, the ECE is given by:
\begin{equation} \label{ece}
    ECE = \sum_{m=1}^{M}\frac{|B_m|}{n}\bigg|acc(B_m)-conf(B_m)\bigg|, \\ 
\end{equation}
where $B_m$ is the set of indices of samples whose prediction confidence falls within interval $I_m = ( \frac{m - 1}{M}, \frac{m}{M} ]$, $|B_m|$ represents the number of predictions in the $m_{th}$ bin, $n$ denotes the total number of samples, and $acc$ and $conf$ denote the average accuracy and average confidence of each bin, respectively.

\section{The Proof of Corollary \ref{cor:corollary}} \label{sec:proof}

\textbf{Theorem 1.} \emph{Let, $x$ is a unlabeled sample under the assumption that the densities $D$ of the observations of for each class $y_1$, and $y_2$ are known.
The Fisher information, $I_{u}(\hat{p})$, for unlabeled samples at the estimate $\hat{p}$ is clearly a measure of the overlap between class conditional densities which denote the information content of unlabeled samples.}

\begin{equation*}
    I_{u}(\hat{p}) = \int\frac{(D(x|y_1) - D(x|y_2))^2}{\hat{p}D(x|y_1) + (1-\hat{p})D(x|y_2)}dx
\end{equation*}

\textbf{Corollary 1.} \emph{The information content of the unlabeled samples that lie in high-density regions ($I^{high}_{u}(\hat{p})$) is greater than those that lie in low-density regions ($I^{low}_{u}(\hat{p})$), i.e., $I^{high}_{u}(\hat{p}) > I^{low}_{u}(\hat{p})$.}

\textbf{\emph{Proof.}} Assume that the dataset satisfies the cluster assumption, which is typical in many semi-supervised learning settings.
Consequently, the following axioms are established:

\textbf{Axiom C.1.} \emph{Data samples in high-density regions satisfy $D(x|y_1) \gg D(x|y_2) \approx 0$ or vice versa.}

\textbf{Axiom C.2.} \emph{Data samples in low-density regions have low values close to 0 for both $D(x|y_1)$ and $D(x|y_2)$, resulting in $D(x|y_1) - D(x|y_2) \approx 0$.}

Let, $\theta_{high}$ be a threshold to filter out unlabeled samples that lie in high-density regions and  $\theta_{low}$ be another threshold to filter out unlabeled samples that lie in low-density regions.
The Fisher information of the unlabeled samples that lie in high-density regions $I^{high}_{u}(\hat{p})$ is:

\begin{equation}
\begin{aligned}
    I^{high}_{u}(\hat{p})&= \int \mathbf{1}_{\{D(x|y_1) > \theta_{high}\}} \frac{(D(x|y_1) - D(x|y_2))^2}{\hat{p}D(x|y_1) + (1-\hat{p})D(x|y_2)} dx
    + 
    \int \mathbf{1}_{\{D(x|y_2) > \theta_{high}\}} \frac{(D(x|y_1) - D(x|y_2))^2}{\hat{p}D(x|y_1) + (1-\hat{p})D(x|y_2)} dx
    \\ 
    &\approx \int \mathbf{1}_{\{D(x|y_1) > \theta_{high}\}} \frac{(D(x|y_1) - 0)^2}{\hat{p}D(x|y_1) + (1-\hat{p})\times0} dx
    + 
    \int \mathbf{1}_{\{D(x|y_2) > \theta_{high}\}} \frac{(0 - D(x|y_2))^2}{\hat{p}\times0 + (1-\hat{p})D(x|y_2)}
    dx  
    \\
    &= \int \mathbf{1}_{\{D(x|y_1) > \theta_{high}\}} \frac{D(x|y_1)}{\hat{p}} dx
    + 
    \int \mathbf{1}_{\{D(x|y_2) > \theta_{high}\}} \frac{D(x|y_2)}{1-\hat{p}}dx 
\end{aligned}
\tag*{By \textbf{Axiom D.1.}}
\end{equation}

By similar reasoning, the Fisher information of the unlabeled sample that lies in low-density regions $I^{low}_{u}(\hat{p})$ is:
\begin{equation}
\begin{aligned}
I^{low}_{u}(\hat{p}) &= \int \mathbf{1}_{\{D(x|y_1) < \theta_{low} \, \text{and} \, D(x|y_2) < \theta_{low}\}} \frac{(D(x|y_1) - D(x|y_2))^2}{\hat{p}D(x|y_1) + (1-\hat{p})D(x|y_2)}dx 
\\
&\approx \int \mathbf{1}_{\{D(x|y_1) < \theta_{low} \, \text{and} \, D(x|y_2) < \theta_{low}\}} \frac{(0)^2}{\hat{p}D(x|y_1) + (1-\hat{p})D(x|y_2)}dx = 0.
\end{aligned}
\tag*{By \textbf{Axiom D.2.}}
\end{equation}
Therefore, the information content of the unlabeled samples that lie in high-density regions is greater than that lie in low-density regions, clearly $I^{high}_{u}(\hat{p}) > I^{low}_{u}(\hat{p})$.

\clearpage

\section{Implementation Details of Prior Knowledge} \label{sec:prior_knowledge_details}
In this section, we describe the implementations of our two density estimators for extracting prior knowledge, the multivariate kernel density estimator and the empirical likelihood.
For both implementations, we utilize \texttt{BorutaShap} \cite{boruta_shap} as the feature selection method.
\subsection{Multivariate Kernel Density Estimator}
To estimate density to regularize the classifier's confidence, we employ a multivariate kernel density estimator provided by the \texttt{statsmodels} package \citep{statsmodels}.
We follow the default kernel settings of \texttt{statsmodels}, which are the Gaussian kernel for continuous features and Aitchison-Aitken kernel for categorical features.
\subsection{Empirical Likelihood}
The empirical likelihood does not require any assumption that the data come from a known family of distributions.
Given the potential for many real-world datasets to be incomplete, distorted, or subject to sampling bias, traditional density estimators might occasionally fall short in approximating true densities. 
Empirical likelihood, with its adaptability, has demonstrated effectiveness in such scenarios, as evidenced by numerous studies \citep{empirical_likelihood_1, empirical_likelihood_2, empirical_likelihood_3}.
Therefore, we adopt empirical likelihood as another measure of density.

We implement a simplified variant of the empirical likelihood as follows:
Let a given sample $\mathbf{x} = \begin{bmatrix}x_{1}, x_{2}, ... , x_{m - 1}, x_{m}\end{bmatrix}$ comprising $m$ features. 
Similarly, a label $\mathbf{y} = \begin{bmatrix}y_{1}, y_{2}, ... , y_{N - 1}, y_{N}\end{bmatrix}$ for the $\mathbf{x}$ is a one-hot vector in the $N$-class dataset.
If $y_{j} = 1$, the sample belongs to the $j^{th}$ class.
For the given $\mathbf{x}$, the empirical likelihood of its pseudo-label $\Tilde{y}_{j}$ is formulated as follows.
\begin{equation} \label{emirical_likelihood}
    \begin{aligned}
    P(\mathbf{x}|\Tilde{y}_{j} = 1) 
        &= \frac{P(\Tilde{y}_{j} = 1 , x_{1}, x_{2}, ... , x_{m - 1}, x_{m})}{P(\Tilde{y}_{j} = 1)} \\
        &= \prod_{k = 1}^{m}P(x_{k}|\Tilde{y}_{j} = 1)
    \end{aligned}
\end{equation}

For simplicity, we operate under the premise that features of $\mathbf{x}$ are conditional independence, given the pseudo-label $\Tilde{y}_{j}$. 
While conditional independence of feature is seldom a reality in many datasets, we are inspired by the assumption used in many successful studies that have used Naive Bayes \footnote{For example, even with correlated features, Naive Bayes, which operates under the conditional independence assumption, often has produced commendable results on a variety of tabular datasets \citep{idiot_nb}}. 
Furthermore, according to Hall's work \citeyearpar{correlation-based_feature_selection}, Good feature subsets contain features highly correlated with the class, yet uncorrelated with each other.
Additionally, there have been some successful research on improving Naive Bayes using feature selection \citep{feature_selection_nb, variable_selection}.
Since we calculate the likelihood of pseudo-labels between selected features, the violation of the assumption is relaxed.
Lastly, we use a log-likelihood by applying logarithms on eq (\ref{emirical_likelihood}) to enhance computational efficiency and prevent numerical errors.

For the categorical features, we determine the likelihood of each distinct value using their empirical distribution. 
For continuous features, we transform them into 10 discrete bins and subsequently calculate their likelihood based on the empirical distribution of these bins.

\clearpage

\section{Statistical Analysis for Empirical Results} \label{subsec:statistical_analysis}

\begin{figure}[!h]
  \centering
  \begin{minipage}{1.0\textwidth}
    \centering
    \begin{minipage}{0.45\linewidth}
    \small
    \setlength{\tabcolsep}{1.5mm}
    \centering
      \begin{tabular}{lccccccr}
        \toprule
        & \multicolumn{2}{c}{FPL} & \multicolumn{2}{c}{CPL} \\
            & statistic & p-value & statistic & p-value \\
        \midrule
        Friedman & 40.1455 & 4.26e-07 & 45.9213 & 3.06e-08 \\
        \midrule
            & \multicolumn{4}{c}{Holm's test with adjusted $\alpha$'s (0.05)} \\
            & \multicolumn{2}{c}{p-values for FPL} & \multicolumn{2}{c}{p-values for CPL}\\
            
         \midrule
        TS  &      \multicolumn{2}{c}{1.0000 } & \multicolumn{2}{c}{1.0000 }\\
        HB  &      \multicolumn{2}{c}{1.0000 } & \multicolumn{2}{c}{1.0000 }\\
        SP  &      \multicolumn{2}{c}{1.0000 } & \multicolumn{2}{c}{1.0000 }\\
        GP  &      \multicolumn{2}{c}{0.3760 } & \multicolumn{2}{c}{0.8486 }\\
        \rowcolor{Gray}
        CAST-D  &      \multicolumn{2}{c}{\textbf{0.0103} } & \multicolumn{2}{c}{\textbf{0.0103} }\\
        \rowcolor{Gray}
        CAST-L  &      \multicolumn{2}{c}{\textbf{0.0103} } & \multicolumn{2}{c}{\textbf{0.0103} }\\
        \bottomrule
      \end{tabular}
      \captionof{table}{Statistic analysis for Table \ref{table:performance_comparison}. Abbreviations are as follows: temperature scaling (TS), histogram binning (HB), spline calibration (SP), and latent Gaussian process (GP).}
       \label{table:statistic}
       \end{minipage}
       \begin{minipage}{0.45\linewidth}
        \centering
        \vspace{4em}
        \begin{subfigure}[b]{1.0\linewidth}
            \centering
            \includegraphics[width=.9\linewidth, height=0.065\textheight]{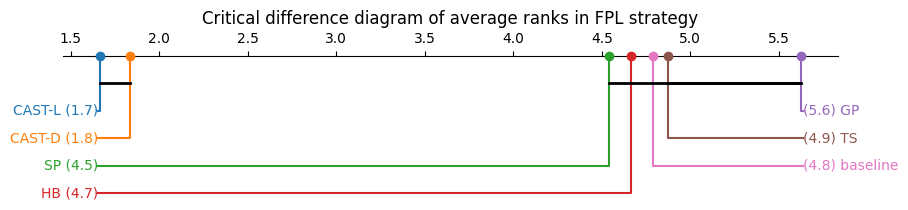}
        \end{subfigure}
        \\
        \begin{subfigure}[b]{1.0\linewidth}
            \centering
            \includegraphics[width=.9\linewidth, height=0.065\textheight]{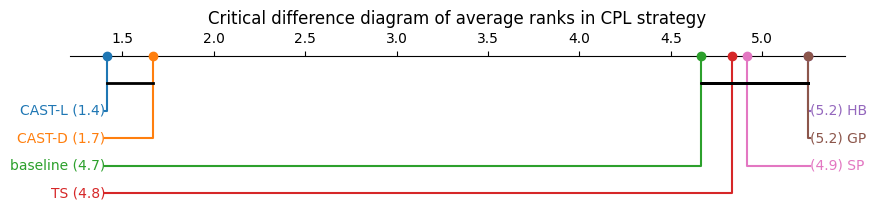}
        \end{subfigure}%
        \vspace{2.8em}
        \captionof{figure}{Critical difference diagrams of average ranks from Table \ref{table:performance_comparison} for FPL (Top) and for CPL (Bottom). Statistically equivalent methods are connected using horizontal bars. Abbreviations are as follows: temperature scaling (TS), histogram binning (HB), spline calibration (SP), and latent Gaussian process (GP).}
        \label{fig:critical_diff}
        \end{minipage}
    \end{minipage}
\end{figure}

To verify the superiority of CAST, we conduct several statistical analyses for Table \ref{table:performance_comparison}.
Firstly, we perform the Friedman test, and the results in Table  \ref{table:statistic} show that we can confidently reject the null hypothesis given the considerably small p-value.
Therefore, we conduct the Conover post-hoc test using the average ranks of each confidence-based self-training method and visualize the results using the critical difference diagrams \cite{critical_diff} shown in Figure \ref{fig:critical_diff}.
We use a significance level $\alpha = 0.05$ for our critical difference diagram.
As shown in Figure \ref{fig:critical_diff}, the regularized confidences of CAST differ significantly from naive confidence at the 95\% confidence level in the self-training contexts, while calibrated confidences using existing confidence calibration methods do not.
We also conduct Holm's test as an additional post-hoc test and report the results between naive confidence and the others in Table \ref{table:statistic}.
This further proves that CAST is significantly different from naive confidence-based self-training.

\vskip -0.4in

\section{How to select the hyperparameter \boldmath$\alpha$?} \label{sec:how_to_select_hyperparameter}
\begin{figure}[htbp]
    \centering
    \includegraphics[width=0.4\linewidth]{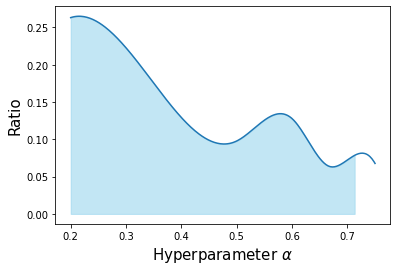}
    \caption{
    The ratio of winning values of the hyperparameter $\alpha$. 
    The colored region denotes the 90\% of confidence interval.} \label{fig:alpha_kde}
\end{figure}

As shown in Section \ref{subsec:ablation_study}, the hyperparameter $\alpha$ plays a crucial role in CAST.
In this section, we study how to select hyperparameter $\alpha$.
Therefore, we analyze the winning value of the hyperparameter $\alpha$ during the grid search for the experiments that are conducted for Table \ref{table:performance_comparison}.
Figure \ref{fig:alpha_kde} depicts a plot summarizing the winning values of $\alpha$.
The $\alpha$ is employed to determine the extent of the influence that prior knowledge on pseudo-label valuation in eq (\ref{eq3}).
Given that prior knowledge sourced from the training data distribution and the confidence of the classifier vary across datasets, models, and random seeds, a universal optimal value does not exist.
However, we can recommend a search range for tuning the $\alpha$.
We identify an upper bound of the 90\% confidence interval for $\alpha$ as 0.7.
Therefore, we suggest 0.7 or less when tuning the hyperparameter $\alpha$.

\clearpage

\section{Additional Experiments}\label{sec:additional_experiments}

\subsection{Computational Cost of CAST} \label{subsec:computational_time}
Here, we demonstrate the negligible computational cost of CAST.
CAST-D has quadratic computational complexity ($O(n^2)$), while CAST-L has linear complexity ($O(n)$) to estimate the density, where $n$ is the number of samples.
We conduct an empirical evaluation of the computational cost of CAST during self-training, utilizing the curriculum pseudo-labeling approach, across four tabular datasets and three commonly used tabular models.
The experiments were carried out using a single CPU core of a Ryzen 5975wx and an RTX 4090 GPU.
Each reported computational cost represents the average across 10 runs.
As shown in Table \ref{table:cost}, the computation times for both CAST-D and CAST-L are comparable to the training time of XGBoost and are significantly lower than those for training neural networks.
These findings underscore that CAST can enhance the performance of existing algorithms with only a marginal increase in computational demands.

\begin{table*}[htbp]
    \centering
    \small
    \setlength{\tabcolsep}{1.5mm}
    \begin{tabular}{lcccccccccccc}
    \toprule
    \multicolumn{13}{c}{Time (s)} \\
    & \multicolumn{3}{c}{6M mortality} & \multicolumn{3}{c}{diabetes} & \multicolumn{3}{c}{ozone} & \multicolumn{3}{c}{cmc} \\
    \cmidrule(r){2-4}
    \cmidrule(r){5-7}
    \cmidrule(r){8-10}
    \cmidrule(r){11-13}
     & XGB & FT & MLP & XGB & FT & MLP & XGB & FT & MLP & XGB & FT & MLP \\
    \midrule
    Training & 5.25 & 749.02 & 96.97 & 0.10 & 26.85 & 9.40 & 0.51 & 89.57 & 14.32 & 0.19 & 27.90 & 16.12 \\
    \midrule
    CAST-D & \multicolumn{3}{c}{7.13} & \multicolumn{3}{c}{0.15} & \multicolumn{3}{c}{0.50} & \multicolumn{3}{c}{0.25}\\
    CAST-L & \multicolumn{3}{c}{3.99} & \multicolumn{3}{c}{0.14} & \multicolumn{3}{c}{0.48} & \multicolumn{3}{c}{0.19}\\
    \midrule
    \multicolumn{13}{c}{Relative additional overhead of CAST compared to training time (\%)} \\
    & \multicolumn{3}{c}{6M mortality} & \multicolumn{3}{c}{diabetes} & \multicolumn{3}{c}{ozone} & \multicolumn{3}{c}{cmc} \\
    \cmidrule(r){2-4}
    \cmidrule(r){5-7}
    \cmidrule(r){8-10}
    \cmidrule(r){11-13}
     & XGB & FT & MLP & XGB & FT & MLP & XGB & FT & MLP & XGB & FT & MLP \\
     \midrule
    CAST-D & 135.84 & 0.95 & 7.35 & 148.15 & 0.57 & 1.63 & 98.72 & 0.56 & 3.51 & 127.70 & 0.89 & 1.53 \\
    CAST-L & 76.08 & 0.53 & 4.12 & 133.65 & 0.52 & 1.47 & 94.99 & 0.54 & 3.38 & 97.35 & 0.68 & 1.17 \\
    \bottomrule
    \end{tabular}
    \caption{Additional computational cost of CAST.}
    \label{table:cost}
\end{table*}

\clearpage

\subsection{Combination of CAST and Noise Filtering} \label{subsec:noise_filtering}

CAST is designed to be seamlessly integrated into existing self-training algorithms without requiring major alterations, making it a versatile add-on. 
This adaptability allows it to be combined with noise filtering techniques for even more effective self-training.
Table \ref{table:CAST_with_noise_filtering} shows the performance improvements when combining CAST with a Mahalanobis distance-based noise filtering approach, as employed by Tanha et al. \citeyearpar{distance_editing}.
Our experimental setup mirrors the one used for Table \ref{table:performance_comparison}, except for the ozone dataset. 
This is because of the challenge of computing the Mahalanobis distance using only 10\% of the labeled data of the ozone dataset.
From the results in Table \ref{table:CAST_with_noise_filtering}, it is clear that the noise filtering approach with CAST provides a greater performance gain.
\begin{table*}[htbp]
  \small
  \centering
  \setlength{\tabcolsep}{1.5mm}
  \begin{tabular}{lccccccccccccc}
    \toprule
     & & \multicolumn{3}{c}{6M mortality} & \multicolumn{3}{c}{diabetes} & \multicolumn{3}{c}{cmc} \\
    \cmidrule(r){3-5}
    \cmidrule(r){6-8}
    \cmidrule(r){9-11}
    \cmidrule(r){12-14}
               &  & XGB   &    FT & MLP & XGB & FT & MLP & XGB & FT & MLP & \\
    \midrule
    \multicolumn{2}{c}{w/o ST} & 0.4055  &    0.3806   &    0.3311   &    0.6613   &    0.7143   &    0.7152   &    0.4638   &    0.4696   &    0.4437 \\
    \midrule
    \multirow{3}{*}{FPL} & Baseline &  \underline{0.4324} & 0.4066 & 0.3635 & 0.6671 & 0.7214 & 0.7249 & 0.4742 & 0.4744 & 0.4440 \\
    & \cellcolor{Gray} CAST-D & \cellcolor{Gray} 0.4297 & \cellcolor{Gray}  \underline{0.4139} & \cellcolor{Gray} \underline{0.3800} & \cellcolor{Gray} \underline{0.6842} & \cellcolor{Gray} \underline{0.7284} & \cellcolor{Gray} \underline{0.7327} & \cellcolor{Gray} \underline{0.4785} & \cellcolor{Gray} \textbf{0.4832} & \cellcolor{Gray} \textbf{0.4559} \\
    & \cellcolor{Gray} CAST-L & \cellcolor{Gray} \textbf{0.4547} & \cellcolor{Gray} \textbf{0.4325} & \cellcolor{Gray} \textbf{0.3982} & \cellcolor{Gray} \textbf{0.6857} & \cellcolor{Gray} \textbf{0.7299} & \cellcolor{Gray} \textbf{0.7336} & \cellcolor{Gray} \textbf{0.4796} & \cellcolor{Gray} \underline{0.4805} & \cellcolor{Gray} \underline{0.4500} \\
    \midrule
    \multirow{3}{*}{CPL} & Baseline &  0.4230 & 0.4128 & 0.3631 & 0.6723 & 0.7288 & 0.7305 & 0.4826 & 0.4921 & 0.4566 \\
    & \cellcolor{Gray} CAST-D   & \cellcolor{Gray}  \underline{0.4255} & \cellcolor{Gray} \underline{0.4211} & \cellcolor{Gray} \underline{0.3737} & \cellcolor{Gray} \textbf{0.6864} & \cellcolor{Gray} \underline{0.7400} & \cellcolor{Gray} \underline{0.7424} & \cellcolor{Gray} \underline{0.4891} & \cellcolor{Gray} \underline{0.5012} & \cellcolor{Gray} \textbf{0.4712} \\
    & \cellcolor{Gray} CAST-L   & \cellcolor{Gray} \textbf{0.4581} & \cellcolor{Gray} \textbf{0.4381} & \cellcolor{Gray} \textbf{0.4009} & \cellcolor{Gray} \underline{0.6829} & \cellcolor{Gray} \textbf{0.7403} & \cellcolor{Gray} \textbf{0.7450} & \cellcolor{Gray} \textbf{0.4895} & \cellcolor{Gray} \textbf{0.5020} & \cellcolor{Gray} \underline{0.4675} \\
    \bottomrule
  \end{tabular}
  \caption{Evaluation results of CAST with Mahalanobis distance-based noise filtering. ``w/o ST" denotes ``without self-training" and the best results are highlighted in bold while the second-best scores are underlined.}
  \label{table:CAST_with_noise_filtering}
\end{table*}

\subsection{CAST Can Change The Most Confident Class}\label{subsec:ability}
Unlike most previous methods regarding reliable pseudo-labeling \cite{setred, uncertainty_pseudo-label, debiased}, CAST can change the most confident class. 
In essence, CAST regularizes the confidence of each pseudo-label based on class-specific prior knowledge. 
Consequently, the most confident class may change because the degree of regularization varies across the classes.
We present results from naive self-training, which strictly determines pseudo-labels based on the most confident class, irrespective of the confidence magnitude \citep{pseudo_label}. 
The results in Table \ref{table:navie_pseudo_labeling} indicate that CAST can modify the most confident class to generate trustworthy pseudo-labels, thereby delivering superior performance over naive confidence-based self-training.
Moreover, this capability explains the results in Section \ref{subsec:noise_filtering}, as many noise filtering techniques identify noise based on the most confident class of unlabeled data.
\begin{table*}[h]
  \setlength{\tabcolsep}{1.5mm}
  \small
  \centering
  \begin{tabular}{lcccccccccccc}
    \toprule
     & \multicolumn{3}{c}{6M mortality} & \multicolumn{3}{c}{diabetes} & \multicolumn{3}{c}{ozone} & \multicolumn{3}{c}{cmc} \\
    \cmidrule(r){2-4}
    \cmidrule(r){5-7}
    \cmidrule(r){8-10}
    \cmidrule(r){11-13}
            &      XGB   &    FT & MLP & XGB & FT & MLP & XGB & FT & MLP & XGB & FT & MLP \\
    \midrule
    w/o ST & 0.4055  &    0.3806   &    0.3311   &    0.6613   &    0.7143   &    0.7152   &    0.2803   &    0.3769   &    0.3823   &    0.4638   &    0.4696   &    0.4437 \\
    \midrule
    Baseline      & \underline{0.4249} & 0.3889 & 0.3501 & 0.6615 & 0.7173 & 0.7204 & 0.2912 & 0.3833 & 0.3879 & 0.4675 & 0.4712 & 0.4441  \\
    \rowcolor{Gray}
    CAST-D & \underline{0.4249} & \underline{0.3897} & \underline{0.3520} & \textbf{0.6639} & \textbf{0.7229} & \textbf{0.7258} & \underline{0.3079} & \textbf{0.3982} & \underline{0.4073} & \underline{0.4716} & \underline{0.4717} & \textbf{0.4457}  \\
    \rowcolor{Gray}
    CAST-L & \textbf{0.4447} & \textbf{0.4042} & \textbf{0.3680} & \underline{0.6621} & \underline{0.7197} & \underline{0.7253} & \textbf{0.3124} & \underline{0.3928} & \textbf{0.4103} & \textbf{0.4743} & \textbf{0.4761} & \underline{0.4442}  \\
    \bottomrule
  \end{tabular}
  \caption{Performance comparison of different confidences in naive self-training. ``w/o ST" denotes ``without self-training" and the best results are highlighted in bold while the second-best scores are underlined.}
  \label{table:navie_pseudo_labeling}
\end{table*}

\clearpage

\subsection{Additional Empirical Results} \label{subsec:additional results}

We conduct extended empirical experiments using an additional seventeen datasets from OpenML-CC18 to demonstrate the superiority of CAST in broader domains including financial, science, bioinformatics, and more.
The experiments are conducted using the same setups which used for Table \ref{table:performance_comparison}.
As shown in Table \ref{table:performance_comparison2}, CAST consistently shows significant improvements in self-training contexts across various tabular domains.

\begin{table}[!h]
  \centering
  \small
  \setlength{\tabcolsep}{1.5mm}
  \begin{tabular}{lcccccccccc}
    \toprule
        &     & kr-vs-kp & credit-g & bank & splice & vehicle & pc4 & pc3 & jm1 & bioresponse \\
    \midrule
    \multicolumn{2}{c}{w/o ST} & 0.9453 & 0.5775 & 0.5526 & 0.9077 & 0.5704 & 0.2467 & 0.2512 & 0.3694 & 0.7436 \\
    \midrule
    \multirow{7}{*}{FPL} & Baseline & 0.9466 & 0.5777 & 0.5613 & \underline{0.9176} & 0.5727 & 0.2719 & 0.2649 & 0.3808 & 0.7434 \\
    & TS & 0.9455 & 0.5774 & 0.5526 & \underline{0.9176} & 0.5727 & 0.2859 & 0.2731 & 0.3712 & 0.7440 \\
    & HB & 0.9462 & 0.5748 & \textbf{0.5671} & 0.9170 & 0.5735 & 0.2549 & 0.2561 & 0.3694 & 0.7438 \\
    & SP & 0.9476 & 0.5777 & 0.5618 & 0.9174 & 0.5722 & 0.2632 & 0.2543 & 0.3759 & 0.7432 \\
    & GP & \textbf{0.9508} & 0.5750 & 0.5536 & 0.9064 & 0.5704 & 0.2593 & 0.2537 & 0.3724 & 0.7478 \\
    \rowcolor{Gray}
    & CAST-D & \underline{0.9499} & \underline{0.5876} & 0.5619 & 0.9122 & \underline{0.5806} & \underline{0.2974} & \underline{0.2808} & \textbf{0.3884} & \underline{0.7487} \\
    \rowcolor{Gray}
    & CAST-L & 0.9496 & \textbf{0.5877} & \underline{0.5628} & \textbf{0.9181} & \textbf{0.5845} & \textbf{0.3004} & \textbf{0.2811} & \underline{0.3834} & \textbf{0.7522} \\
    \midrule
    \multirow{7}{*}{CPL} & Baseline & 0.9485 & 0.5827 & 0.5568 & 0.9113 & 0.5737 & 0.2650 & 0.2695 & \underline{0.3883} & 0.7448 \\
    & TS & 0.9492 & 0.5752 & 0.5590 & 0.9108 & 0.5737 & 0.2912 & 0.2615 & 0.3816 & 0.7458 \\
    & HB & 0.9491 & 0.5768 & 0.5617 & \textbf{0.9153} & 0.5802 & 0.2837 & 0.2585 & 0.3712 & 0.7462 \\
    & SP & 0.9490 & 0.5794 & 0.5556 & 0.9119 & 0.5708 & 0.2620 & 0.2668 & 0.3850 & 0.7456 \\
    & GP & 0.9509 & 0.5710 & 0.5533 & 0.9126 & \textbf{0.5935} & 0.2609 & 0.2574 & 0.3864 & 0.7455 \\
    \rowcolor{Gray}
    & CAST-D & \underline{0.9518} & \underline{0.5931} & \underline{0.5626} & 0.9132 & 0.5755 & \textbf{0.3103} & \underline{0.2849} & \textbf{0.3885} & \textbf{0.7619} \\
    \rowcolor{Gray}
    & CAST-L & \textbf{0.9527} & \textbf{0.5991} & \textbf{0.5640} & \underline{0.9133} & \underline{0.5876} & \underline{0.3014} & \textbf{0.2918} & 0.3872 & \underline{0.7572} \\
    \midrule
    &     & kc2 & adult & blood & qsar-biodeg & robot & churn & car & steel & Avg Rank \tiny{(std)} \\
    \midrule
    \multicolumn{2}{c}{w/o ST} & 0.5091 & 0.8187 & 0.7302 & 0.7291 & 0.9624 & 0.5859 & 0.5472 & 0.6762 & - \\
    \midrule
    \multirow{7}{*}{FPL} & Baseline & 0.5182 & 0.8209 & 0.7342 & 0.7287 & 0.9633 & 0.6051 & 0.5512 & 0.6748 & 4.26\tiny{(1.14)} \\
    & TS & 0.5109 & 0.8204 & 0.7307 & 0.7321 & 0.9633 & 0.5913 & 0.5512 & 0.6748 & 4.97\tiny{(1.40)} \\
    & HB & 0.5105 & \textbf{0.8228} & 0.7302 & 0.7359 & 0.9635 & 0.5926 & 0.5476 & \underline{0.6793} & 4.65\tiny{(1.97)} \\
    & SP & 0.5181 & 0.8205 & 0.7416 & 0.7271 & 0.9639 & 0.6059 & 0.5471 & 0.6756 & 4.82\tiny{(1.29)} \\
    & GP & 0.5084 & 0.8187 & 0.7431 & 0.7279 & 0.9624 & 0.5899 & 0.5474 & 0.6762 & 5.56\tiny{(1.78)} \\
    \rowcolor{Gray}
    & CAST-D & \textbf{0.5367} & 0.8208 & \textbf{0.7438} & \textbf{0.7406} & \textbf{0.9658} & \underline{0.6204} & \textbf{0.5718} & 0.6762 & \underline{2.03\tiny{(1.22)}} \\
    \rowcolor{Gray}
    & CAST-L & \underline{0.5243} & \underline{0.8224} & \underline{0.7436} & \underline{0.7378} & \underline{0.9653} & \textbf{0.6207} & \underline{0.5637} & \textbf{0.6856} & \textbf{1.71\tiny{(0.82)}} \\
    \midrule
    \multirow{7}{*}{CPL} & Baseline & 0.5226 & 0.8187 & 0.7427 & 0.7278 & 0.9630 & 0.6133 & 0.5631 & 0.6764 & 5.06\tiny{(1.61)} \\
    & TS & 0.5124 & \textbf{0.8241} & 0.7298 & 0.7350 & 0.9630 & 0.5934 & 0.5631 & 0.6764 & 5.12 \tiny{(1.69)} \\
    & HB & 0.5260 & \underline{0.8237} & 0.7400 & \underline{0.7386} & 0.9648 & 0.5957 & 0.5603 & 0.6785 & 4.06\tiny{(1.70)} \\
    & SP & 0.5221 & 0.8190 & 0.7400 & 0.7281 & 0.9631 & 0.6095 & 0.5626 & 0.6774 & 5.24\tiny{(0.081)} \\
    & GP & 0.5217 & 0.8204 & 0.7431 & 0.7309 & \underline{0.9659} & 0.6163 & 0.5548 & \underline{0.6805} & 4.65\tiny{(2.00)} \\
    \rowcolor{Gray}
    & CAST-D & \textbf{0.5553} & 0.8219 & \textbf{0.7509} & 0.7378 & \textbf{0.9660} & \textbf{0.6379} & \textbf{0.5768} & 0.6767 & \underline{2.00\tiny{(1.19)}} \\
    \rowcolor{Gray}
    & CAST-L & \underline{0.5269} & 0.8227 & \underline{0.7460} & \textbf{0.7401} & 0.9656 & \underline{0.6327} & \underline{0.5693} & \textbf{0.6863} & \textbf{1.88\tiny{(0.83)}} \\
    \bottomrule
  \end{tabular}
  \caption{Performance comparison over seventeen tabular datasets. The best results are highlighted in bold, while the second-best scores are underlined. Abbreviations are as follows: without self-training (w/o ST), temperature scaling (TS), histogram binning (HB), spline calibration (SP), and latent Gaussian process (GP).}
  \label{table:performance_comparison2}
\end{table}

\clearpage

\section{Additional Experimental Details} \label{sec:experimental_details}
\subsection{Details of Datasets} \label{subsec:dataset_details}
\begin{table}[htbp]
  \centering
  \small
  \setlength{\tabcolsep}{1.5mm}
  \begin{tabular}{lcccclcccc}
    \toprule
    name & class & features & n\_samples & metric & name & class & features & n\_samples & metric \\
    \midrule
    6M mortality & 2 & 76 & 15628 & F1 & jm1 & 2 & 22 & 10855 & F1 \\
    diabetes & 2 & 9 & 768 & acc & bioresponse & 2 & 1777 & 3751 & F1 \\
    ozone (ozone-level-8hr) & 2 & 73 & 2534 & F1 & kc2 & 2 & 22 & 522 & F1 \\
    cmc & 3 & 10 & 1473 & b-acc & adult & 2 & 15 & 48842 & b-acc \\
    kr-vs-kp & 2 & 37 & 3196 & acc & blood-transfusion-service-center & 2 & 5 & 748 & acc \\
    credit-g & 2 & 21 & 1000 & b-acc & qsar-biodeg & 2 & 42 & 1055 & b-acc \\
    bank (bank-marketing) & 2 & 17 & 45211 & f1 & wall-robot-navigation & 4 & 25 & 5456 & F1 \\
    splice & 3 & 62 & 3190 & b-acc & churn & 2 & 21 & 5000 & F1 \\
    vehicle & 4 & 19 & 846 & acc & car & 4 & 7 & 1728 & b-acc \\
    pc4 & 2 & 38 & 1458 & F1 & steel-plates-fault & 7 & 28 & 1941 & F1 \\
    pc3 & 2 & 38 & 1563 & F1 & & & & & \\
    \bottomrule
  \end{tabular}
  \caption{Overview of datasets. We abbreviate ``F1-score" as ``F1," ``balanced accuracy" as ``b-acc," and ``accuracy" as ``acc".}
  \label{table:details_dataset}
\end{table}
\textbf{Dataset preprocessing.}
We use label encoding for all categorical features, except for the 6M mortality dataset where certain categorical features necessitate one-hot encoding.
We impute missing data using an iterative imputer in \texttt{scikit-learn} package \citep{sklearn}.
For MLP, we embed categorical features in high-dimensional spaces and apply batch normalization to the continuous features. 
\subsection{Implementation Details of Tabular Models}
We use Pytorch Tabular framework for FT-Transformer and MLP \citep{pytorch_tabular}, and the official Python package for XGBoost.
VIME, SubTab, SCARF, and SwitchTab are implemented.
While the official implementation of VIME uses only one reconstruction loss, we use two reconstruction losses for categorical features and continuous features, respectively.
This results in the use of two hyperparameters, $\alpha_1$ and $\alpha_2$, for the reconstruction losses, whereas the official implementation only uses one hyperparameter $\alpha$.

\subsection{Details of Confidence Calibration}
We use netcal framework \citep{netcal} for temperature scaling and histogram binning.
We adopt six knots for the spline calibration according to the Gupta el al.'s \citeyearpar{spline_calibration} work and do not use any hyperparameters for the latent Gaussian process calibration since it is a nonparametric method.
Then, we fit calibration methods to the validation dataset based on the ECE score, except for spline calibration, which does not require a fitting procedure.
\subsection{Details of Hyperparameter Tunning}
\begin{table}[htbp]
\begin{center}
\setlength{\tabcolsep}{1.5mm}
  \small
\begin{tabular}{lcccr}
\toprule
Hyperparameter & Search Method & Search Space \\
\midrule
max\_leaves & suggest\_int & [300,4000] \\
n\_estimators & suggest\_int & [10,3000] \\
learning\_rate & suggest\_uniform & [0,1] \\
max\_depth & suggest\_int & [3, 20] \\
scale\_pos\_weight & suggest\_int & [1, 100] \\
\bottomrule
\end{tabular}
\end{center}
\caption{Optuna hyperparameter search space for XGBoost}
\label{xgboost_search_space}
\end{table}

\begin{table}[htbp]
\begin{center}
\setlength{\tabcolsep}{1.5mm}
  \small
\begin{tabular}{lcccr}
\toprule
Hyperparameter & Search Method & Search Space \\
\midrule
input\_embed\_dim & suggest\_categorical & [16,24,32,48] \\
embedding\_dropout & suggest\_uniform & [0.05,0.3] \\
share\_embedding & suggest\_categorical & [True, False] \\
num\_heads & suggest\_categorical & [1,2,4,8] \\
num\_attn\_blocks & suggest\_int & [2,10] \\
transformer\_activation & suggest\_categorical & [GEGLU, ReGLU, SwiGLU] \\
use\_batch\_norm & suggest\_categorical & [True, False] \\
batch\_norm\_continuous\_input & suggest\_categorical & [True, False] \\
learning\_rate & suggest\_uniform & [0.0001, 0.05] \\
scheduler\_gamma & suggest\_uniform & [0.1, 0.95] \\
scheduler\_step\_size & suggest\_int & [10, 100] \\ 
\bottomrule
\end{tabular}
\end{center}
\caption{Optuna hyperparameter search space for FT-Transformer}
\label{fttransformer_search_space}
\end{table}

\begin{table*}[tbp]
\begin{center}
\setlength{\tabcolsep}{1.5mm}
  \small
\begin{tabular}{lcccr}
\toprule
Hyperparameter & Search Method & Search Space \\
\midrule
embedding\_dropout & suggest\_uniform & [0, 0.2] \\
layers & suggest\_categorical & [128-64-32, 256-128-64, 128-64-32-16, 256-128-64-32] \\
activation & suggest\_categorical & [ReLU, LeakyReLU] \\
learning\_rate & suggest\_uniform & [0.0001, 0.05] \\
scheduler\_gamma & suggest\_uniform & [0.1, 0.95] \\
scheduler\_step\_size & suggest\_int & [10, 100] \\
\bottomrule
\end{tabular}
\end{center}
\caption{Optuna hyperparameter search space for MLP}
\label{mlp_search_space}
\end{table*}

\begin{table*}[tbp]
\begin{center}
\setlength{\tabcolsep}{1.5mm}
  \small
\begin{tabular}{lcccr}
\toprule
Hyperparameter & Search Method & Search Space \\
\midrule
predictor\_hidden\_dim & suggest\_int & [16, 512] \\
$p_m$ & suggest\_float & [0.1, 0.9] \\
$\alpha_1$ & suggest\_float & [0.1, 5] \\
$\alpha_2$ & suggest\_float & [0.1, 5] \\
beta & suggest\_float & [0.1, 10] \\
K & suggest\_int & [2, 20] \\
learning\_rate & suggest\_uniform & [0.0001, 0.05] \\
scheduler\_gamma & suggest\_uniform & [0.1, 0.95] \\
scheduler\_step\_size & suggest\_int & [10, 100] \\
\bottomrule
\end{tabular}
\end{center}
\caption{Optuna hyperparameter search space for VIME}
\label{vime_search_space}
\end{table*}

\begin{table*}[tbp]
\begin{center}
\setlength{\tabcolsep}{1.5mm}
  \small
\begin{tabular}{lcccr}
\toprule
Hyperparameter & Search Method & Search Space \\
\midrule
emb\_dim & suggest\_int & [4, 1024] \\
$\tau$ & suggest\_float & [0.05, 0.15] \\
use\_cosine\_similarity & suggest\_categorical & [True, False] \\
use\_contrastive & suggest\_categorical & [True, False] \\
use\_distance & suggest\_categorical & [True, False] \\
n\_subsets & suggest\_int & [2, 7] \\
overlap\_ratio & suggest\_float & [0, 1] \\
mask\_ratio & suggest\_float & [0.1, 0.3] \\
noise\_level & suggest\_float & [0.5, 2] \\
noise\_type & suggest\_categorical & [Swap, Gaussian, Zero\_Out] \\
learning\_rate & suggest\_uniform & [0.0001, 0.05] \\
scheduler\_gamma & suggest\_uniform & [0.1, 0.95] \\
scheduler\_step\_size & suggest\_int & [10, 100] \\
\bottomrule
\end{tabular}
\end{center}
\caption{Optuna hyperparameter search space for SubTab}
\label{subtab_search_space}
\end{table*}

\begin{table*}[t]
\begin{center}
\setlength{\tabcolsep}{1.5mm}
  \small
\begin{tabular}{lcccr}
\toprule
Hyperparameter & Search Method & Search Space \\
\midrule
emb\_dim & suggest\_int & [16, 512] \\
encoder\_depth & suggest\_int & [2, 6] \\
head\_depth & suggest\_int & [1, 3] \\
corruption\_rate & suggest\_float & [0, 0.7] \\
dropout\_rate & suggest\_float & [0.05, 0.3] \\
learning\_rate & suggest\_uniform & [0.0001, 0.05] \\
scheduler\_gamma & suggest\_uniform & [0.1, 0.95] \\
scheduler\_step\_size & suggest\_int & [10, 100] \\
\bottomrule
\end{tabular}
\end{center}
\caption{Optuna hyperparameter search space for SCARF}
\label{scarf_search_space}
\end{table*}

\begin{table*}[t]
\begin{center}
\setlength{\tabcolsep}{1.5mm}
  \small
\begin{tabular}{lcccr}
\toprule
Hyperparameter & Search Method & Search Space \\
\midrule
encoder\_head\_dim & suggest\_int & [16, 256] \\
encoder\_depth & suggest\_int & [1, 4] \\
n\_head & suggest\_int & [1, 4] \\
ffn\_factor & suggest\_int & [1, 3] \\
dropout\_rate & suggest\_float & [0.05, 0.3] \\
learning\_rate & suggest\_float & [0.0001, 0.05] \\
weight\_decay & suggest\_float & [1e-5, 5e-4] \\
\bottomrule
\end{tabular}
\end{center}
\caption{Optuna hyperparameter search space for SwitchTab}
\label{switchtab_search_space}
\end{table*}

\end{appendices}

\end{document}